%% file: eccv2020arxiv.tex
\documentclass[runningheads]{llncs}
\usepackage{graphicx}
\usepackage{tikz}
\usepackage{comment}
\usepackage{amsmath,amssymb}
\usepackage{color}
\usepackage[normalem]{ulem}
\usepackage{multirow}
\usepackage{import}
\usepackage{makecell}
\usepackage{caption}
\usepackage{subcaption}
\usepackage[ruled,vlined]{algorithm2e}

\usepackage{amsmath}
\usepackage{amssymb}
\usepackage{booktabs}

\usepackage{arydshln}

\usepackage[pagebackref=true,breaklinks=true,letterpaper=true,colorlinks,bookmarks=false]{hyperref}
\usepackage{kotex}
\usepackage{graphbox}
\graphicspath{{assets/}}

\newcommand\eg{\textit{e.g.}}
\newcommand\ie{\textit{i.e.}}

\def\fullmodel{Dual Memory-augmented Font Generation Network}
\def\model{DM-Font}

\newcommand\emd{EMD~\cite{zhang2018_cvpr_emd}}
\newcommand\funit{FUNIT~\cite{liu2019funit}}
\newcommand\agis{AGIS-Net~\cite{gao2019agisnet}}
\newcommand\ours{\model{} (ours)}

\usepackage[width=122mm,left=12mm,paperwidth=146mm,height=193mm,top=12mm,paperheight=217mm]{geometry}

\begin{document}
\pagestyle{headings}
\mainmatter

\title{Few-shot Compositional Font Generation \\ with Dual Memory}

\titlerunning{Few-shot Compositional Font Generation with Dual Memory}
\author{Junbum Cha \and
Sanghyuk Chun \and
Gayoung Lee \and \\
Bado Lee \and
Seonghyeon Kim \and
Hwalsuk Lee}
\authorrunning{J. Cha, S. Chun, G. Lee, B. Lee, S. Kim, H. Lee}
\institute{Clova AI Research, NAVER Corp.\\
\email{\{junbum.cha, sanghyuk.c, gayoung.lee, \\bado.lee, kim.seonghyeon, hwalsuk.lee\}@navercorp.com}
}

\maketitle

\begin{abstract}
Generating a new font library is a very labor-intensive and time-consuming job for glyph-rich scripts. Despite the remarkable success of existing font generation methods, they have significant drawbacks; they require a large number of reference images to generate a new font set, or they fail to capture detailed styles with only a few samples. In this paper, we focus on compositional scripts, a widely used letter system in the world, where each glyph can be decomposed by several components. By utilizing the compositionality of compositional scripts, we propose a novel font generation framework, named \fullmodel{} (\model{}), which enables us to generate a high-quality font library with only a few samples. We employ memory components and global-context awareness in the generator to take advantage of the compositionality. In the experiments on Korean-handwriting fonts and Thai-printing fonts, we observe that our method generates a significantly better quality of samples with faithful stylization compared to the state-of-the-art generation methods quantitatively and qualitatively. Source code is available at \url{https://github.com/clovaai/dmfont}.
\end{abstract}

\begin{figure}[t]
    \centering
    \includegraphics[width=.9\linewidth]{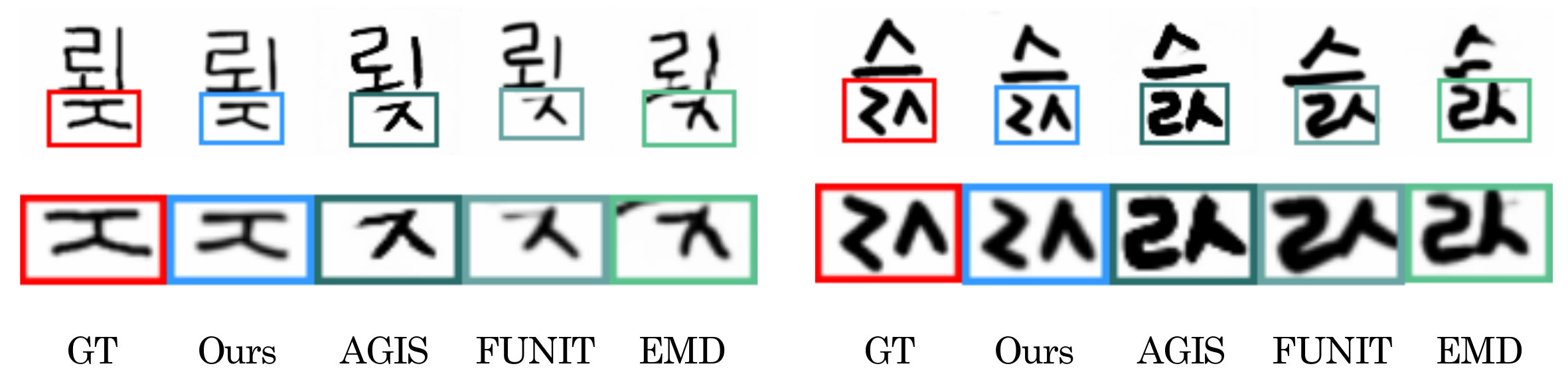}
    \caption{{\bf Few-shot font generation results.} While previous few-shot font generation methods (AGIS~\cite{gao2019agisnet}, FUNIT~\cite{liu2019funit}, and EMD~\cite{zhang2018_cvpr_emd}) are failed to generate unseen font, our model successfully transfer the font style and details.}
    \label{fig:teaser}
\end{figure}

\section{Introduction}
Advances of web technology lead people to consume a massive amount of texts on the web. It makes designing a new font style, \eg, personalized handwriting, critical. However, because traditional methods to make a font library heavily rely on expert designers by manually designing each glyph, creating a font library is extremely expensive and labor-intensive for glyph-rich scripts such as Chinese (more than 50,000 glyphs), Korean (11,172 glyphs), or Thai (11,088 glyphs)~\cite{koreantextbook}.

Recently, end-to-end font generation methods~\cite{zi2zi,jiang2017dcfont,jiang2019_aaai_scfont,lyu2017_icdar_aegg,chang2018_bmvc_hgan,chang2018_wacv_densecyclegan} have been proposed to build a font set without human experts. The methods solve image-to-image translation tasks between various font styles based on generative adversarial networks (GANs)~\cite{gan}. While the methods have shown the remarkable achievement, they still require a large number of samples, \eg, $775$ samples~\cite{jiang2017dcfont,jiang2019_aaai_scfont} to generate a new font set. Moreover, they require additional training to create a new glyph set, \ie, they need to finetune the pretrained model on the given new glyph subset. Thus, these finetune-based methods are rarely practical if collecting the target glyphs is extremely expensive, \eg, human handwriting.

Several recent studies attempt to generate a font set without additional training with a large number of glyphs, but using only a few samples~\cite{azadi2018mcgan,sun2018_ijcai_savae,zhang2018_cvpr_emd,gao2019agisnet,srivatsan2019_emnlp_deepfactorization}. Despite their successful few-shot generation performances on training styles, existing few-shot font generation methods often fail to generate high-quality font library with unseen style few-shot samples as illustrated in Figure~\ref{fig:teaser}. We solve this problem using the inherent glyph characteristics in contrast to most of the previous works handling the problem in the end-to-end data-driven manner without any human prior. A few researchers have considered characteristics of glyphs to improve font generation methods~\cite{sun2018_ijcai_savae,jiang2019_aaai_scfont}, but their approaches are either still requiring more than $700$ samples~\cite{jiang2019_aaai_scfont}, or only designed for memory efficiency~\cite{sun2018_ijcai_savae}.

In this paper, we focus on a famous family of scripts, called {\it compositional scripts}, which are composed of a combination of sub-glyphs or components. For example, the Korean script has 11,172 valid glyphs with only 68 components. One can build a full font library by designing only 68 sub-glyphs and combine them by the pre-defined rule. However, this rule-based method has a significant limitation; a sub-glyph changes its shape and position diversely depending on the combination, as shown in Figure~\ref{fig:compositional-script}. Hence, even if a user has a complete sub-glyphs, generating a full font set is impossible without the combination rule of components. Due to the limitations, compositional scripts have been manually designed for each glyph despite its compositionality~\cite{koreantextbook}.

Our framework for the few-shot font generation, \fullmodel{} (\model{}), utilizes the compositionality supervision in the weakly-supervised manner, \ie, no component-wise bounding box or mask is required but only component labels are required, resulting on more efficient and effective generation. We employ the dual memory structure ({\it persistent memory} and {\it dynamic memory}) to efficiently capture the global glyph structure and the local component-wise styles, respectively. This strategy enables us to generate a new high-quality font library with only a few samples, \eg, $28$ samples and $44$ samples for Korean and Thai, respectively. In the experiments, the generated Korean and Thai fonts show both quantitatively better visual quality in various metrics and qualitatively being preferred in the user study.

\section{Related Works}
\subsection{Few-shot image-to-image translation}
Image-to-image (I2I) translation~\cite{isola2017_cvpr_pix2pix,zhu2017_iccv_cyclegan,stargan,karras2019stylegan,starganv2} aims to learn the mapping between different domains. This mapping preserves the content in the source domain while changing the style as the target domain. Mainstream I2I translation methods assume an abundance of target training samples which is impractical. To deal with more realistic scenarios where the target samples are scarce, few-shot I2I translation works appeared recently~\cite{liu2019funit}. These methods can be directly applied to the font generation task as a translation task between the reference font and target font. We compare our method with FUNIT~\cite{liu2019funit}.

As an independent line of research, style transfer methods~\cite{gatys2016neuralstyle,wct,adain,deepphotostyle,photowct,wct2} have been proposed to transfer styles of an unseen reference while preserving the original content. Unlike I2I translation tasks, style transfer methods cannot be directly transformed to font generation tasks, because they usually define the style as the set of textures and colors. However, in font generation tasks, the style of font is usually defined as discriminative local property of the font. Hence, our work does not concern style transfer methods as our baseline.

\subsection{Automatic font generation}
Automatic font generation task is an I2I translation between different font domains, \ie, styles. We categorize the automatic font generation methods into two classes, which are many-shot and few-shot methods, according to way to generate a new font set. Many-shot methods~\cite{zi2zi,jiang2017dcfont,lyu2017_icdar_aegg,chang2018_bmvc_hgan,chang2018_wacv_densecyclegan,jiang2019_aaai_scfont} directly finetune the model on the target font set with a large number of samples, \eg, $775$. It is impractical in many real-world scenarios when collecting new glyphs is costly, \eg, handwriting.

In contrast, few-shot font generation methods~\cite{zhang2018_cvpr_emd,srivatsan2019_emnlp_deepfactorization,azadi2018mcgan,gao2019agisnet,sun2018_ijcai_savae} does not require additional finetuning and a large number of reference images. However, the existing few-shot methods have significant drawbacks. For example, some methods generate a whole font set at single forward path~\cite{azadi2018mcgan,srivatsan2019_emnlp_deepfactorization}. Hence, they require a huge model capacity and cannot be applied to glyph-rich scripts but scripts with only a few glyphs, \eg, Latin alphabet. 
On the other hand, EMD~\cite{zhang2018_cvpr_emd} and AGIS-Net~\cite{gao2019agisnet} can be applied to any general scripts, but they show worse synthesizing quality to unseen style fonts, as observed in our experimental results.
SA-VAE~\cite{sun2018_ijcai_savae}, a Chinese-specific method, keeps the model size small by compressing one-hot character-wise embeddings based on compositionality of Chinese script. Compared with SA-VAE, ours handles the features as component-wise, not character-wise. It brings huge advantages in not only reducing feature dimension but also in performances as shown in our experimental results.

\begin{figure}[t]
    \centering
    \includegraphics[width=0.8\linewidth]{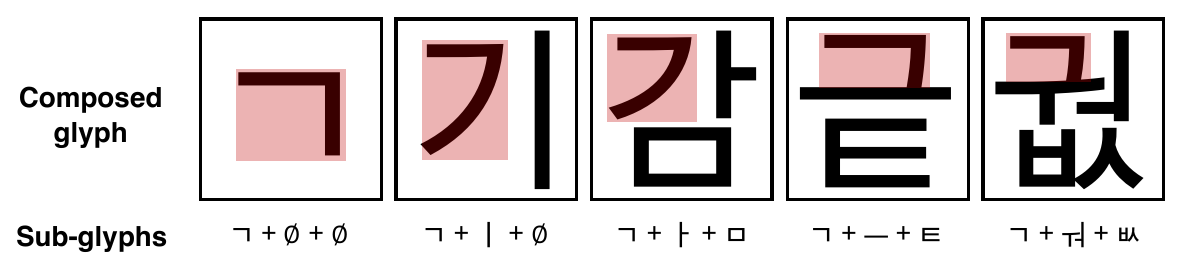}
    \caption{{\bf Examples of compositionality of Korean script.} Even if we choose the same sub-glyph, \eg, ``ㄱ'', the shape and position of each sub-glyph are varying depending on the combination, as shown in red boxes.}
    \label{fig:compositional-script}
\end{figure}

\section{Preliminary: Complete Compositional Scripts}

{\it Compositional script} is a widely-used glyph-rich script, where each glyph can be decomposed by several components as shown in Fig.~\ref{fig:compositional-script}. These scripts account for 24 of the top 30 popular scripts, including Chinese, Hindi, Arabic, Korean, Thai. A compositional script is either {\it complete} or not, where each glyph in {\it complete compositional scripts} can be decomposed to fixed number sub-glyphs. For example, every Korean glyph can be decomposed by three sub-glyphs (See Fig.~\ref{fig:compositional-script}). Similarly, a Thai character has four components. Furthermore, complete compositional letters have specific sub-glyph sets for each {\it component type}. For example, the Korean alphabet has three component types where each component type has $19$, $21$, $28$ sub-glyphs. By combining them, Korean letter has $19 \times 21 \times 28 = 11,172$ valid characters. Note that the minimum number of glyphs to get the entire sub-glyph set is $28$. Similarly, Thai letter can represent $44 \times 7 \times 9 \times 4 = 11,088$ characters, and $44$ characters are required to cover whole sub-glyphs.

Some compositional scripts are not complete. For example, each character of the Chinese letter can be decomposed into a diverse number of sub-glyphs. Although we mainly validate our method on Korean and Thai scripts, our method can be easily extended to other compositional scripts.

\begin{figure}[t]
\centering
\begin{subfigure}[h]{0.9\textwidth}
\includegraphics[width=\textwidth]{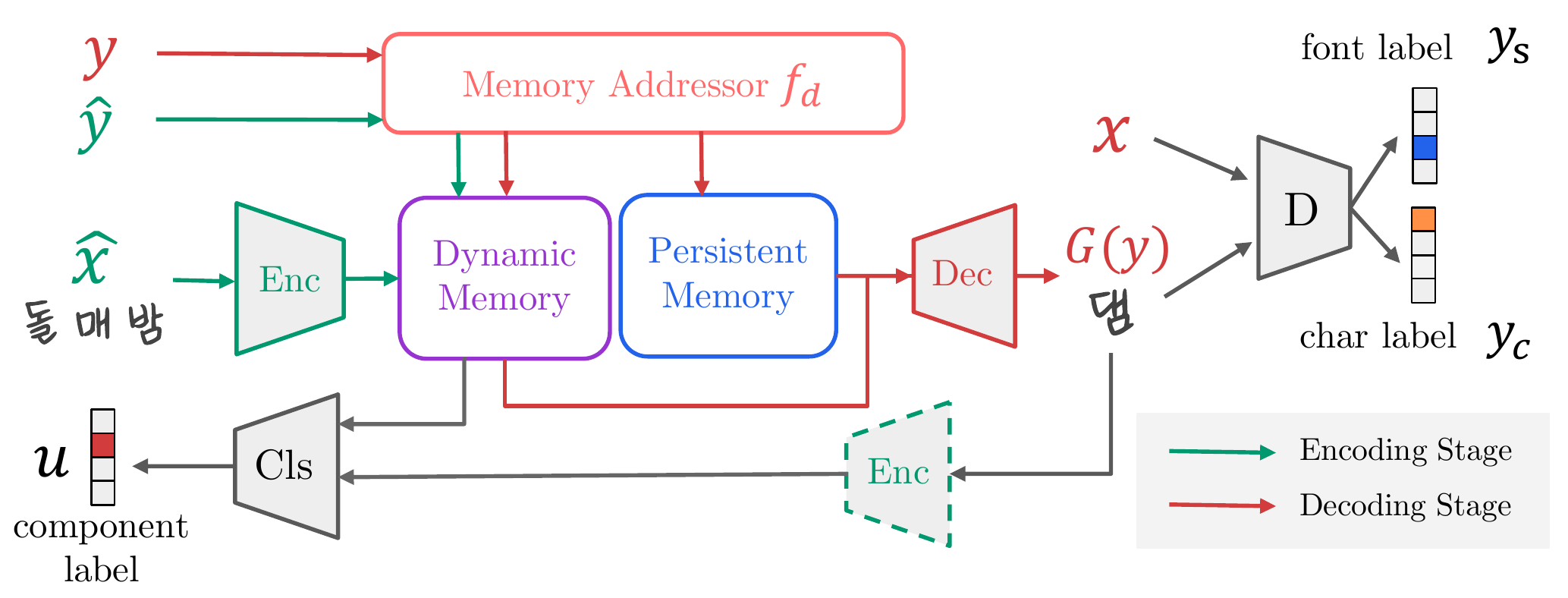}
\caption{Architecture overview.}\label{fig:arch_all}
\end{subfigure}
\begin{subfigure}[h]{0.45\textwidth}
\includegraphics[page=1, width=\textwidth]{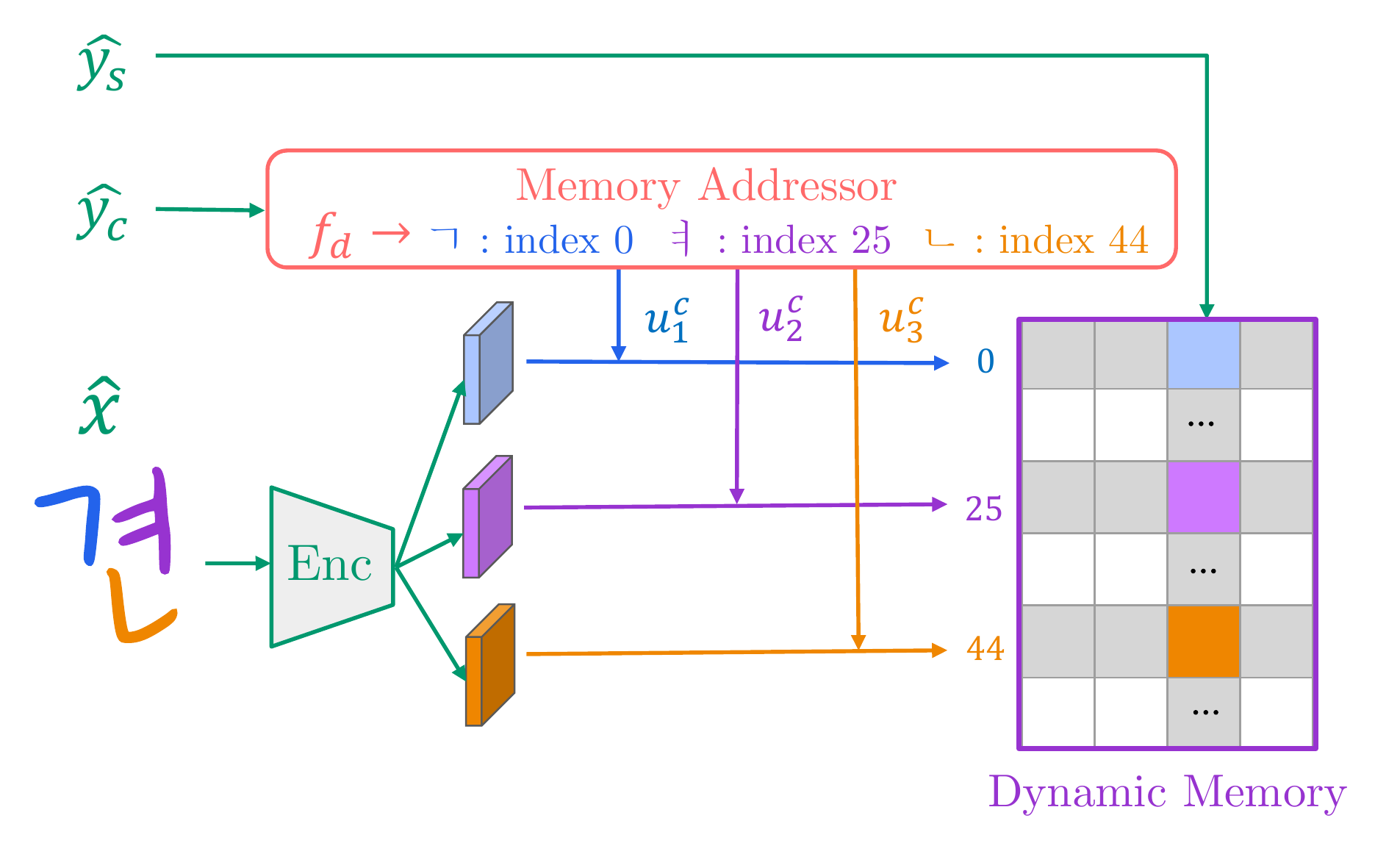}
\caption{Encoding phase detail.}\label{fig:memory_encode}
\end{subfigure}%
\begin{subfigure}[h]{0.45\textwidth}
\includegraphics[page=2, width=\textwidth]{assets/memory_detail_v3.pdf}
\caption{Decoding phase detail.}\label{fig:memory_decode}
\end{subfigure}
\caption{{\bf \model{} overview.} (a) The model encodes the reference style glyphs and stores the component-wise features into the memory -- (b). The decoder generates images with the component-wise features -- (c). (b) The encoder extracts the component-wise features and stores them into the dynamic memory using the component label $u^c_i$ and the style label $\hat y_s$. (c) The memory addressor loads the component features by the character label $y_c$ and feeds them to the decoder.
}
\label{fig:arch}
\end{figure}

\section{\fullmodel}

In this section, we introduce a novel architecture, \fullmodel{} (\model{}), which utilizes the compositionality of a script by the augmented dual memory structure. \model{} disentangles global composition information and local styles, and writes them into persistent and dynamic memory, respectively. It enables to make a high-quality full glyph library only with very few references, \eg, $28$ samples for Korean, $44$ samples for Thai.

\subsection{Architecture overview}

We illustrate the architecture overview of \model{} in Fig.~\ref{fig:arch_all}. The generation process consists of encoding and decoding stages. In the encoding stage, the reference style glyphs are encoded to the component features and stored into the dynamic memory. After the encoding, the decoder fetches the component features and generates the target glyph according to the target character label.

{\bf Encoder} $Enc$ disassembles a source glyph into the several component features using the pre-defined decomposition function. We adopt multi-head structure, one head per one component type. The encoded component-wise features are written into the dynamic memory as shown in Figure~\ref{fig:memory_encode}.

We employ two memory modules, where {\bf persistent memory} (PM) is a component-wise learned embedding that represents the intrinsic shape of each component and the global information of the script such as the compositionality, while {\bf dynamic memory} (DM) stores encoded component features of the given reference glyphs. Hence, PM captures the global information of sub-glyphs independent to each font style, while encoded features in DM learn unique local styles depending on each font. Note that DM simply stores and retrieves the encoded features, but PM is learned embedding trained from the data. Therefore, DM is adaptive to the reference input style samples, while PM is fixed after training. We provide detailed analysis of each memory in the experiments. 

{\bf Memory addressor} provides the access address of both dynamic and persistent memory based on the given character label $y_c$ as shown in Figure~\ref{fig:memory_encode} and Figure~\ref{fig:memory_decode}. We use pre-defined decomposition function $f_d: y_c \mapsto \{u^c_i ~|~ i = 1 \ldots M_c\}$ to get the component-wise address, where $u^c_i$ is the label of i-th component of $y_c$, and $M_c$ is the number of sub-glyphs for $y_c$. For example, the function decomposes a Korean character, ``한'' by $f_d (\text{``한''}) = $ \{``ㅎ'', ``ㅏ'', ``ㄴ''\}. 
The function maps input character to Unicode and decomposes it by a simple rule.
More details of the decomposition function are given in Appendix~\ref{sec:appendix_arch_detail}.

The component-wise encoded features for the reference $\hat x$, whose character label is $\hat y_c$ and style label is $\hat y_s$, are stored into DM during the encoding stage. In our scenario, the encoder $Enc$ is a multi-head encoder, and $\hat y_c$ can be decomposed by $f_d(\hat y_c)$ to sub-glyph labels $\hat u^c_i$. Hence, the features in DM at address $(\hat u^c_i, \hat y_s)$, $DM(\hat u^c_i, \hat y_s)$ is computed by $Enc_i(\hat x)$, where $i$ is the index of the component type and $Enc_i$ is the encoder output corresponding to $i$.

In the decoding stage, \textbf{decoder} $Dec$ generates a target glyph with the target character $y_c$ and the reference style $y_s$ using the component-wise features stored into the dynamic memory $DM$ and the persistent memory $PM$ as the following:
\begin{equation}\label{eq:decoder}
    G(y_c, y_s) = Dec \big( \big[ DM(u^c_i, y_s),\, PM(u^c_i) ~|~ u^c_i \in f_d(y_c) \big] \big),
\end{equation}
where $[x_0, \ldots, x_n]$ refers to the concatenation operation.

For the better generation quality, we also employ a discriminator and a component classifier. For {\bf discriminator} $D$, we use a multitask discriminator \cite{mescheder2018training,liu2019funit} with the font condition and the character condition. The multitask discriminator has independent branches for each target class and each branch performs binary classification. Considering two types of conditions, we use two multitask discriminator, one for character classes and the other for font classes, with a shared backbone. We further use {\bf component classifier} $Cls$ to ensure the model to fully utilize the compositionality. The component classifier provides additional supervision to the generator that stabilizes the training.

Moreover, we introduce the global-context awareness and local-style preservation to the generator, called \textbf{compositional generator}. Specifically, self-attention blocks \cite{cao2019gcnet,zhang2019sagan} are used in the encoder to facilitate relational reasoning between components, and the hourglass block \cite{newell2016hourglass,Lin_2017_CVPR_FPN} is attached to the decoder to aware global-context while preserving locality. In the experiment section, we analyze the impact of the architectural improvements on the final performance. We provide the architecture and the implementation details in Appendix~\ref{sec:appendix_arch_detail}.

\model{} learns the compositionality in the weakly-supervised manner; it does not require any exact component location, \eg, component-wise bounding boxes, but only component labels are required. Hence, \model{} is not restricted to the font generation only, but can be applied to any generation task with compositionality, \eg, attribute conditioned generation tasks. Extending \model{} to attribute labeled datasets, \eg, CelebA~\cite{celeba}, will be an interesting topic.

\subsection{Learning}

We train \model{} from font sets $(x, y_c, y_f) \sim \mathcal D$, where $x$ is a target glyph image, $y_c$ and $y_f$ is a character and font label, respectively. During the training, we assume that different font labels represent different styles, \ie, we set $y_s = y_f$ in equation~\eqref{eq:decoder}. Also, for the efficiency, we only encode a core component subset to compose the target glyph $x$ into the DM instead of the full component set. For example, the Korean script has the full component set with size $68$, but only $3$ components are required to construct a single character.

We use {\bf adversarial loss} to let the model generate plausible images.
\begin{equation}
\label{eq:adv_obj}
    \mathcal L_{adv} = \mathbb E_{x, y} \left[ \log D_y(x) \right] + \mathbb E_{x, y} \left[ \log (1 - D_y(G(y_c, y_f))) \right],
\end{equation}
where $G$ generates an image $G(y_c, y_f)$ from the given image $x$ and target label $y$ by equation~\eqref{eq:decoder}. The discriminator $D_y$ is conditional on the target label $y$. We employed two types of the discriminator to solve the problem. The font discriminator is a conditional discriminator on the source font index and the character discriminator aims to classify what is the given character.

{\bf $\mathbf{L_1}$ loss} adds supervision from the ground truth target $x$ as the following:
\begin{equation}
    \mathcal L_{l1} = \mathbb E_{x, y} \left[ \| x - G(y_c, y_f) \|_1 \right].
\end{equation}

We also use {\bf feature matching loss} to improve the stability of the training. The feature matching loss is constructed using the output from the $l$-th layer of the $L$-layered discriminator, $D^{(l)}_f$.
\begin{equation}
    \mathcal L_{feat} = \mathbb E_{x, y} \left[ \frac{1}{L}\sum_{l=1}^L\| D_f^{(l)}(x) - D_f^{(l)}( G(y_c, y_f) ) \|_1 \right].
\end{equation}

Lastly, to let the model fully utilize the compositionality, we train the model with additional {\bf component-classification loss}. For the given input $x$, we extract the component-wise features using the encoder $Enc$, and train them with cross-entropy loss (CE) using component labels $u \in f_d(y_c)$, where $f_d$ is the component decomposition function to the given character label $y_c$.

\begin{equation}
    \mathcal L_{cls} = \mathbb E_{x, y} \left[ \sum_{u^c_i \in f_d(y_c)} \text{CE} (Enc_i(x), u^c_i) \right] + \mathbb E_{y} \left[ \sum_{u^c_i \in f_d(y_c)} \text{CE} (Enc_i(G(y_c, y_f)), u^c_i) \right].
\end{equation}

The final objective function to optimize the generator $G$, the discriminator $D$, and the component classifier $C$ is defined as the following:
\begin{equation}
    \min_{G,C} \max_D \mathcal{L}_{adv(font)} + \mathcal{L}_{adv(char)} + \lambda_{l1} \mathcal{L}_{l1} + \lambda_{feat}\mathcal{L}_{feat} + \lambda_{cls}\mathcal{L}_{cls},
\end{equation}
where $\lambda_{l1}, \lambda_{feat}, \lambda_{cls}$ are control parameters to importance of each loss function. We set $\lambda_{l1} = 0.1, \lambda_{feat} = 1.0, \lambda_{cls} = 0.1$ for all experiments.

\section{Experiments}

\subsection{Datasets} \label{sec:datasets}

\subsubsection{Korean-handwriting dataset.} Due to its diversity and data sparsity, generating a handwritten font with only a few samples is challenging. We validate the models using $86$ Korean-handwriting fonts\footnote{We collect public fonts from  \url{http://uhbeefont.com/}.} refined by the expert designer. Each font library contains $2,448$ widely-used Korean glyphs. We train the models using $80\%$ fonts and $90\%$ characters, and validate the models on the remaining split. We separately evaluate the models on the seen ($90\%$) and unseen ($10\%$) characters to measure the generalizability to the unseen characters. $30$ characters are used for the reference.

\subsubsection{Thai-printing dataset.} Compared with Korean letters, Thai letters have more complex structure; Thai characters are composed of four sub-glyphs while Korean characters have three components. We demonstrate the models on $105$ Thai-printing fonts\footnote{\url{https://github.com/jeffmcneill/thai-font-collection}.}. The train-evaluation split strategy is same as Korean-handwriting experiments, and $44$ samples are used for the few-shot generation.

\subsubsection{Korean-unrefined dataset.} We also gather unrefined Korean handwriting dataset from $88$ non-experts, letting each applicant write $150$ characters. This dataset is extremely diverse and not refined by expert designers different from the Korean-handwriting dataset. We use the Korean-unrefined dataset as the validation of the models trained on the Korean-handwriting dataset, \ie, the Korean-unrefined dataset is not visible during the training, but only a few samples are visible for the evaluation. $30$ samples are used for the generation as well as the Korean-handwriting dataset.

\subsection{Comparison methods and evaluation metrics} \label{sec:baselines}
\subsubsection{Comparison methods.} We compare our model with state-of-the-art few-shot font generation methods, including EMD~\cite{zhang2018_cvpr_emd}, AGIS-Net~\cite{gao2019agisnet}, and FUNIT~\cite{liu2019funit}. We exclude the methods which are Chinese-specific~\cite{sun2018_ijcai_savae} or not applicable to glyph-rich scripts~\cite{srivatsan2019_emnlp_deepfactorization}.
Here, we slightly modified FUNIT, originally designed for unsupervised translation, by changing its reconstruction loss to $L_1$ loss with ground truths and conditioning the discriminator to both contents and styles.

\import{tables/}{quantitative-new.tex}

\subsubsection{Evaluation metrics.} \label{sec:eval-metrics}
Assessing a generative model is difficult because of its non-tractability. Several quantitative evaluation metrics~\cite{johnson2016_eccv_perceptual,heusel2017_nips_ttur_fid,zhang2018_cvpr_lpips,ferjad2020ganeval} have attempted to measure the performance of the trained generative model with different assumptions, but it is still controversial what is the best evaluation methods for generative models.
In this paper, we consider three diverse levels of evaluation metrics; pixel-level, perceptual-level and human-level evaluations.

{\bf Pixel-level evaluation metrics} assess the pixel structural similarity between the ground truth image and the generated image. We employ the structural similarity index (SSIM) and multi-scale structural similarity index (MS-SSIM).

However, pixel-level metrics often disagree with human perceptions. Thus, we also evaluate the models with {\bf perceptual-level evaluation metrics}. We trained four ResNet-50~\cite{he2016_cvpr_resnet} models on the Korean-handwriting dataset and Thai-printing dataset to classify style and character label. Unlike the generation task, the whole fonts and characters are used for the training. More detailed classifier training settings are in Appendix~\ref{sec:appendix_impl_detail}. We denote a metric is {\it context-aware} if the metric is performed using the content classifier, and {\it style-aware} is defined similarly. Note that the classifiers are independent to the font generation models, but only used for the evaluation. We report the top-1 accuracy, perceptual distance (PD) \cite{johnson2016_eccv_perceptual,zhang2018_cvpr_lpips}, and mean FID (mFID) \cite{liu2019funit} using the classifiers. PD is computed by $L_2$ distance of the features between generated glyph and GT glyph, and mFID is a conditional FID~\cite{heusel2017_nips_ttur_fid} by averaging FID for each target class.

Finally, we conduct a user study on the Korean-unrefined dataset for measuring {\bf human-level evaluation metric}. We ask users about three types of preference: content preference, style preference, and user preference considering both content and style.
The questionnaire is made of 90 questions, 30 for each preference. Each question shows 40 glyphs, consisting of 32 glyphs generated by four models and 8 GT glyphs. The order of choices is shuffled for anonymity. We collect total 3,420 responses from 38 Korean natives. More details of user study are provided in Appendix~\ref{sec:appendix_impl_detail}.

\begin{figure}[t]
    \centering
    \begin{subfigure}[b]{\textwidth}
        \centering
        \includegraphics[page=3,width=0.99\textwidth]{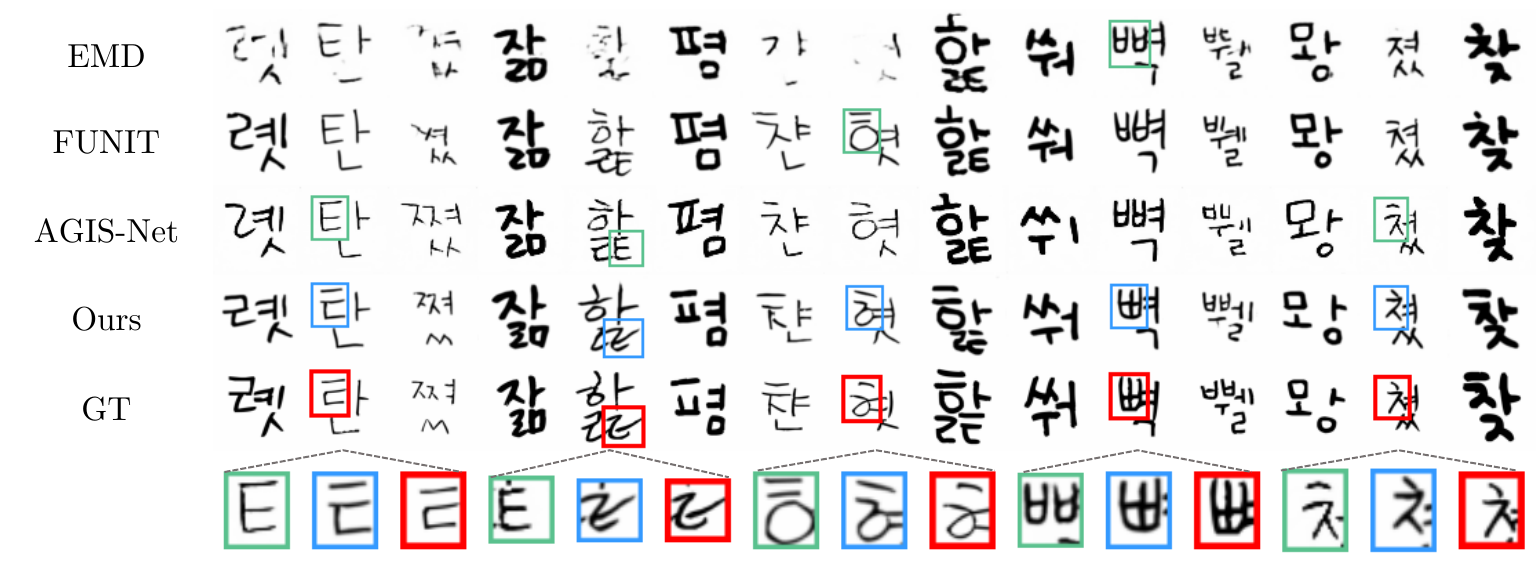}
        \caption{Seen character set during training.}
    \end{subfigure}
    \begin{subfigure}[b]{\textwidth}
        \centering
        \includegraphics[page=4,width=0.99\textwidth]{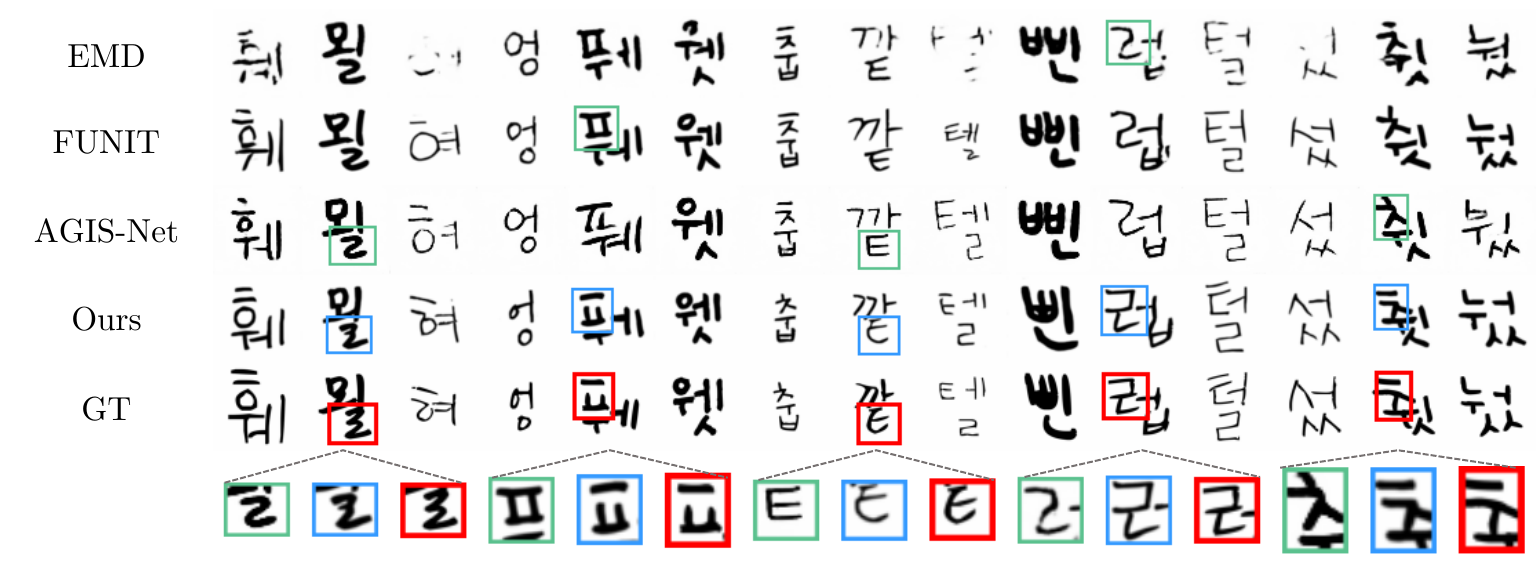}
        \caption{Unseen character set during training.}
    \end{subfigure}
    \caption{{\bf Qualitative comparison on the Korean-handwriting dataset.} Visualization of generated samples with seen and unseen characters. We show insets of baseline results (green box), ours (blue box) and ground truth (red box).
    Ours successfully transfers the detailed style of the target style, while baselines fail to generate glyphs with the detailed reference style.
    }
    \label{fig:qualitative-kor}
\end{figure}

\begin{figure}[t]
    \centering
    \begin{subfigure}[b]{\textwidth}
        \centering
        \includegraphics[page=1,width=0.99\textwidth]{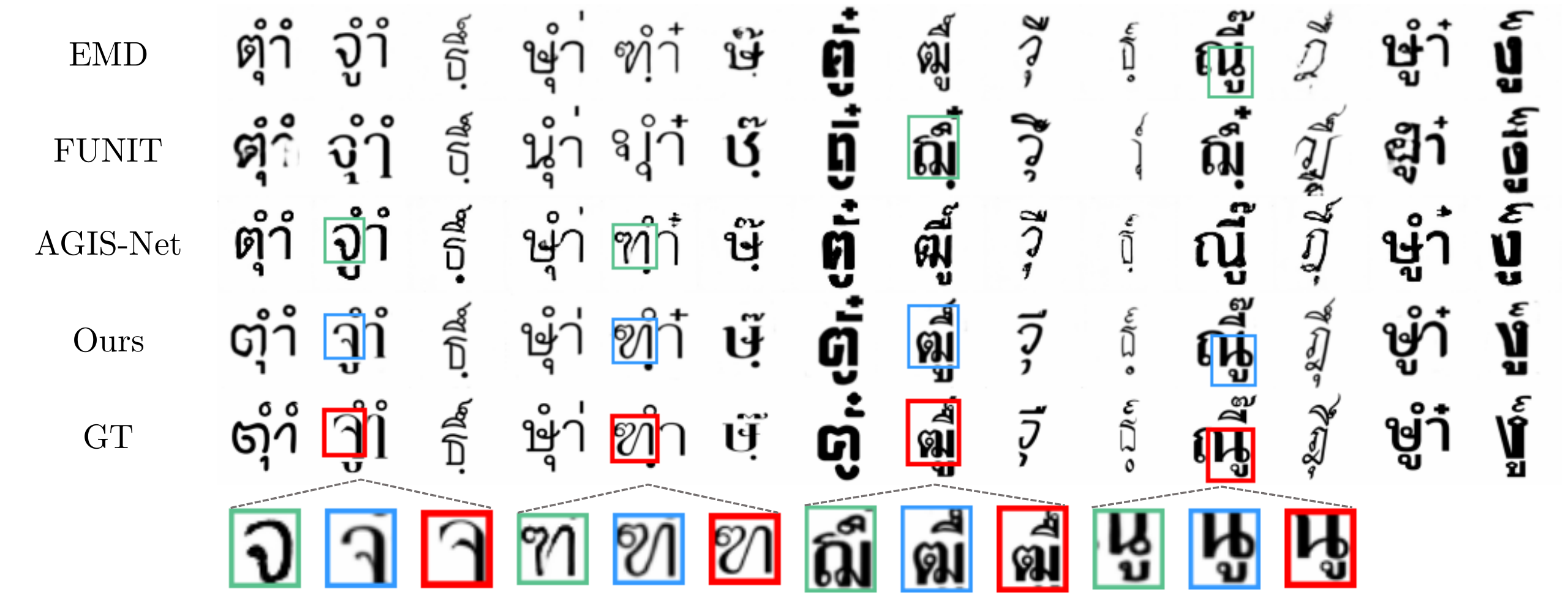}
        \caption{Seen character set during training.}
    \end{subfigure}
    \begin{subfigure}[b]{\textwidth}
        \centering
        \includegraphics[page=2,width=0.99\textwidth]{assets/fig-horizontal-results-thai-inset_v2.pdf}
        \caption{Unseen character set during training.}
    \end{subfigure}
    \caption{{\bf Qualitative comparison on the Thai-printing dataset.} Visualization of generated samples with seen and unseen characters. We show insets of baseline results (green box), ours (blue box) and ground truth (red box).
    Overall, ours faithfully transfer the target style, while other methods even often fail to preserve contents in unseen character sets.
    }
    \label{fig:qualitative-thai}
\end{figure}

\subsection{Main results} \label{sec:main_results}

\subsubsection{Quantitative evaluation.} \label{sec:quantitative_eval}
The main results on Korean-handwriting and Thai-printing datasets are reported in Table~\ref{table:quantitative-kor} and Table~\ref{table:quantitative-thai}, respectively. We also report the evaluation results on the Korean-unrefined dataset in Appendix~\ref{sec:appendix_results}. We follow the dataset split introduced in Section~\ref{sec:datasets}. In the experiments, \model{} remarkably outperforms the comparison methods in most of evaluation metrics, especially on style-aware benchmarks. Baseline methods show slightly worse content-aware performances on unseen characters than seen characters, \eg, AGIS-Net shows worse content-aware accuracy ($98.7 \rightarrow 98.3$), PD ($0.018 \rightarrow 0.019$), and mFID ($23.9 \rightarrow 25.9$) in Table~\ref{table:quantitative-kor}. In contrast, \model{} consistently shows better generalizability to the unobserved characters during the training for both datasets. It is because our model interprets a glyph at the component level, the model easily extrapolates the unseen characters from the learned component-wise features stored in memory modules.

Our method shows significant improvements in style-aware metrics. \model{} achieves $62.6\%$ and $50.6\%$ accuracy while other methods show much less accuracy, \eg, about $5\%$ for Korean unseen and Thai unseen character sets, respectively. Likewise, the model shows dramatic improvements in perceptual distance and mFID as well as the accuracy measure. In the latter section, we provide more detailed analysis that the baseline methods are overfitted to the training styles and failed to generalize to unseen styles.

\subsubsection{Qualitative comparison.} \label{sec:qualitative_eval}
We also provide visual comparisons in Figure~\ref{fig:qualitative-kor} and Figure~\ref{fig:qualitative-thai}, which contain various challenging fonts including thin, thick, and curvy fonts. Our method generates glyphs with consistently better visual quality than the baseline methods. EMD~\cite{zhang2018_cvpr_emd} often erases thin fonts unintentionally, which causes low content scores compared with the other baseline methods. FUNIT~\cite{liu2019funit} and AGIS-Net~\cite{gao2019agisnet} accurately generate the content of glyphs and capture global styles well including overall thickness and font sizes. However, the detailed styles of the components in their results look different from the ground truths. Moreover, some generated glyphs for unseen Thai style lose the original content (see the difference between green boxes and red boxes in Figure~\ref{fig:qualitative-kor} and Figure~\ref{fig:qualitative-thai} for more details). Compared with the baselines, our method generates the most plausible images in terms of global font styles and detailed component styles. These results show that our model preserves details in the components using the dual memory and reuse them to generate a new glyph.

\subsubsection{User study.} \label{sec:user_study}
\begin{figure}
    \centering
    \includegraphics[width=0.9\textwidth]{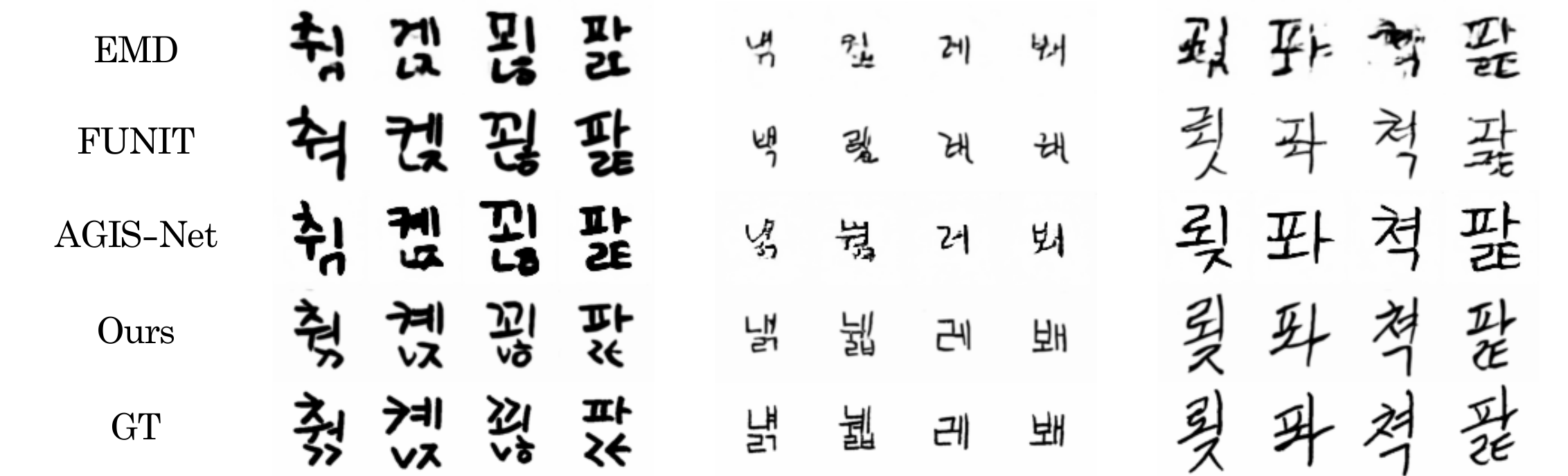}
    \caption{Samples for the user study. The Korean-unrefined dataset is used.}
    \label{fig:user-study}
\end{figure}

We conduct a user study to further evaluate the methods in terms of human preferences using the Korean-unrefined dataset. Example generated glyphs are illustrated in Figure~\ref{fig:user-study}. Users are asked to choose the most preferred generated samples in terms of content preservation, faithfulness to the reference style, and personal preference. The results are shown in Table~\ref{table:user_study}, which present similar intuitions with Table~\ref{table:quantitative-kor}; AGIS-Net and our method are comparable in the content evaluation, and our method is dominant in the style preference.

\import{tables/}{user_study.tex}

\subsection{More analysis} \label{sec:analysis}
\subsubsection{Ablation study.} \label{sec:abalation}
\import{tables/}{ablation.tex}
We investigate the impact of our design choices by ablative studies. Table~\ref{table:ablation-components} shows that the overall performances are improved by adding proposed components such as dynamic memory, persistent memory, and compositional generator. We report full table in Appendix~\ref{sec:appendix_results}.

Here, the baseline method is similar to FUNIT whose content and style accuracies are $93.9$ and $5.4$, respectively. The baseline suffers from the failure of style generalization as previous methods. We observe that dynamic memory and persistent memory dramatically improves style scores while preserving content scores. Finally, our architectural improvements bring the best performance.

We also explore the performance influence of each objective. As shown in Table~\ref{table:ablation-objective}, removing $L_1$ loss and feature matching loss slightly degrades performances. The component-classification loss, which enforces the compositionality to the model, is the most important factor for successful training.

\import{tables/}{vis_nn_l1.tex}

\subsubsection{Style overfitting of baselines.}
We analyze the generated glyphs using our style classifier to investigate the style overfitting of the baseline methods. Figure~\ref{fig:style-memorization} shows the predicted classes for each model output. We observe that the baseline methods often generate samples similar to the training samples. On the other hand, our model avoids the style overfitting by learning the compositionality of glyphs and directly reusing components of inputs. Consequently, as supported by previous quantitative and qualitative evaluations, our model is robust to the out-of-distributed font generation compared to the existing methods. We provide more analysis of the overfitting of comparison methods in the Appendix~\ref{sec:appendix_results}.

\subsubsection{Component-wise style mixing.} \label{sec:component-mixing}
\begin{figure}[t]
    \centering
    \includegraphics[width=.9\textwidth]{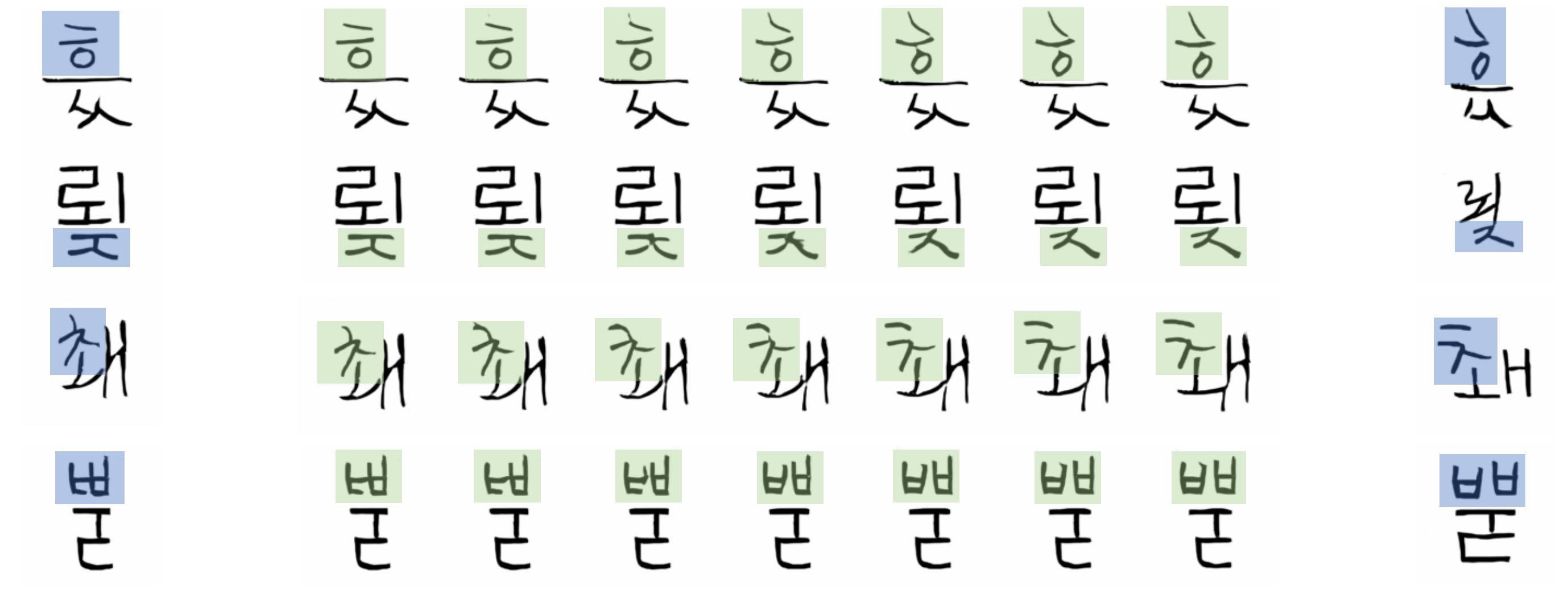}
    \caption{{\bf Component-wise style mixing.} We interpolate only one component (marked by blue boxes) between two glyphs (the first column and the last column). The interpolated sub-glyphs are marked by green boxes. Our model successfully interpolates two sub-glyphs, while preserving other local styles.}
    \label{fig:component-mixing}
\end{figure}

In Figure~\ref{fig:component-mixing}, we demonstrate our model can interpolate styles component-wisely. It supports that our model fully utilizes the compositionality to generate a glyph.

\section{Conclusions}

Previous few-shot font generation methods often fail to generalize to unseen styles. In this paper, we propose a novel few-shot font generation framework for compositional scripts, named \fullmodel{} (\model{}). Our method effectively incorporates the prior knowledge of compositional script into the framework via two external memories: the dynamic memory and the persistent memory. \model{} utilizes the compositionality supervision in the weakly-supervised manner, \ie, neither component-wise bounding box nor mask used during the training. The experimental results showed that the existing methods fail in stylization on unseen fonts, while \model{} remarkably and consistently outperforms the existing few-shot font generation methods on Korean and Thai letters. Extensive empirical evidence support that our framework lets the model fully utilize the compositionality so that the model can produce high-quality samples with only a few samples.

\section*{Acknowledgement}
We thank Clova OCR and Clova AI Research team for discussion and advice, especially Song Park for the internal review.

{\small
\bibliographystyle{splncs04}
\bibliography{egbib}
}

\clearpage
\appendix
\numberwithin{equation}{section}
\numberwithin{figure}{section}
\numberwithin{table}{section}

\section{Network Architecture Details}\label{sec:appendix_arch_detail}
\subsection{Memory addressors}
\begin{algorithm}[H]
\SetAlgoLined
\KwIn{A character label $y_c$}
\KwOut{Component labels $u_1^c$, $u_2^c$, and $u_3^c$}
\KwData{The number of components for each $i$-th component type $N_i$.}
\SetKwFunction{ToUnicode}{ToUnicode}
\SetKwFunction{div}{div}
\SetKwFunction{mod}{mod}
unicode = \ToUnicode{$y_c$} \\
\tcp{0xAC00 is the initial Korean Unicode}
code = unicode - 0xAC00 \\
$u_3^c$ = code \mod $N_3$ \\
$u_2^c$ = (code \div $N_3$) \mod $N_2$\\
$u_1^c$ = code \div ($N_3 \times N_2$)
\label{alg:decompose-kor}
\caption{Unicode-based Korean letter decomposition function}
\end{algorithm}

The memory addressor converts character label $y_c$ to the set of component labels $u^c_i$ by the pre-defined decomposition function $f_d: y_c \mapsto \{u^c_i ~|~ i = 1 \ldots M_c\}$, where $u^c_i$ is the label of i-th component of $y_c$ and $M_c$ is the number of sub-glyphs for $y_c$. In this paper, we employ Unicode-based decomposition functions specified to each language. We describe the decomposition function for Korean script as an example in Algorithm~\ref{alg:decompose-kor}. The function disassembles a character into component labels by the pre-defined rule. On the other hand, each Thai character consists of several Unicodes, each of which corresponds to one component. Therefore, each Unicode constituting the letter is a label itself. The Thai decomposition function only needs to determine the component type of each Unicode.

\subsection{Network architecture}
\begin{figure}[t]
\centering
\includegraphics[page=2,width=\textwidth]{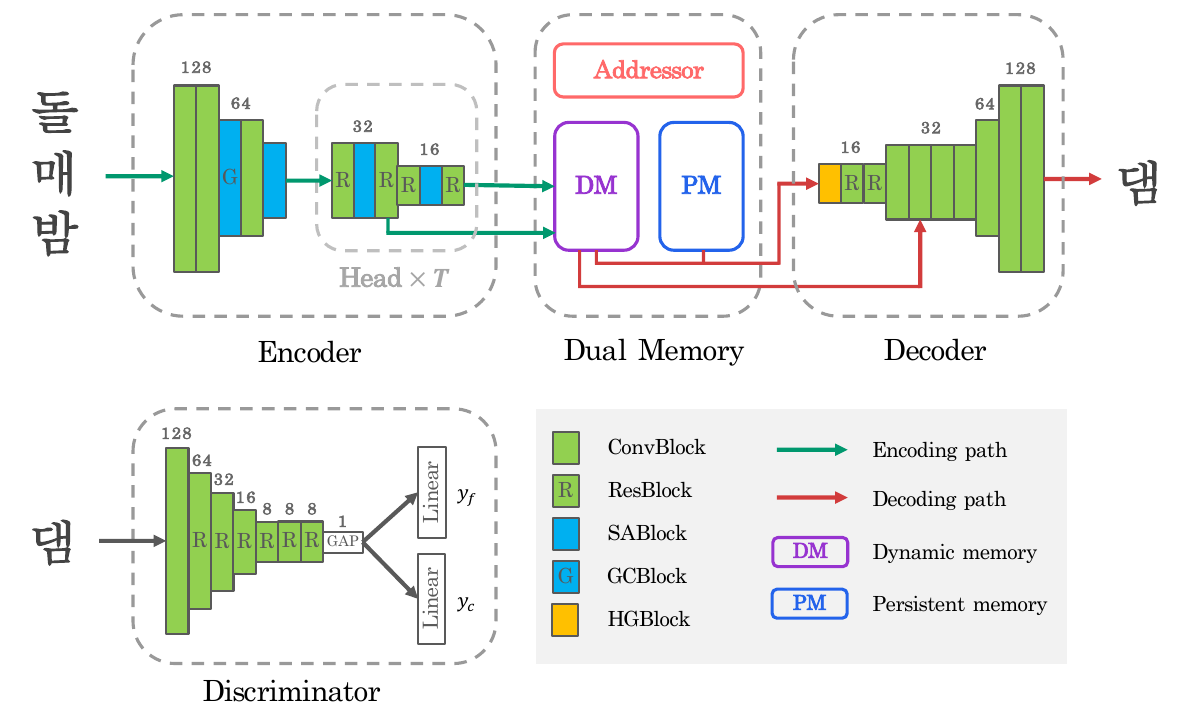}
\caption{The encoder holds multiple heads according to the number of component types $T$. We denote the spatial size of each block in the figure.}
\label{fig:net-arch}
\vspace{-1em}
\end{figure}

The proposed architecture has two important properties: global-context awareness and local-style preservation. Global-context awareness allows the relational reasoning between components to the network, boosting to disassemble source glyphs into sub-glyphs and assemble them to the target glyph. Local-style preservation indicates that the local style of source glyph is reflected in the target.

For the global-context awareness, the encoder adopts global-context block (GCBlock) \cite{cao2019gcnet} and self-attention block (SABlock) \cite{zhang2019sagan,vaswani2017_NIPS_Transformer_Attention}, and the decoder employs hourglass block (HGBlock) \cite{newell2016hourglass,Lin_2017_CVPR_FPN}.
These blocks extend the receptive field globally and facilitate relational reasoning between components while preserving locality. For the local-style preservation, the network handles multi-level features based on the dual memory framework. The specific architecture overview is described visually in Figure \ref{fig:net-arch}.

The generator consists of five modules; convolution block (ConvBlock), residual block (ResBlock), self-attention block, global-context block, and hourglass block. Our SABlock is adopted from Transformer \cite{vaswani2017_NIPS_Transformer_Attention} instead of SAGAN \cite{zhang2019sagan}, \ie, the block consists of multi-head self-attention and position-wise feed-forward. We also use two-dimensional relative positional encoding from \cite{bello2019_ICCV_AttentionAugmentedConv}. The hourglass block consists of multiple convolution blocks and downsampling or upsampling operation follows each block. Through hourglass structure, the spatial size of the feature map is reduced to $1 \times 1$ and restored to the original size, which extends the receptive field globally preserving locality. The channel size starts at 32 and doubles as blocks are added, up to 256 for the encoder and 512 for the decoder.

We employ a simple structure for the discriminator. Several residual blocks follow the first convolution block. Like the generator, the channel size starts at 32 and doubles as blocks are added, up to 1024. The output feature map of the last residual block is spatially squeezed to $1 \times 1$ size and it is fed to the two linear, font and character discriminators. Each discriminator is a multi-task discriminator that performs binary classification for each target class. Therefore, the font discriminator produces $|\mathbb Y_f|$ binary outputs and the character discriminator produces $|\mathbb Y_c|$ binary outputs, where $|\mathbb Y|$ denotes the number of target classes.

Since the persistent memory (PM) is independent of local styles, we set the size of PM same as the size of high-level features, the final output of the encoder, \ie, $16 \times 16$. The learned embedding is refined via three convolution blocks, added to the high-level features of dynamic memory (DM), and then fed to the decoder. The component classifier comprises two residual blocks and one linear layer and identifies the class of the high-level component features from the DM.

\section{Experimental Setting Details}
\label{sec:appendix_impl_detail}
\subsection{\model{} implementation details}

We use Adam \cite{kingma2015adam} with a learning rate of 0.0002 for the generator and 0.0008 for the discriminator, following the two time-scale update rule \cite{heusel2017_nips_ttur_fid}. The component classifier use same learning rate with the generator. The discriminator adopts spectral normalization \cite{miyato2018spectral} for the regularization. We train the model with hinge GAN loss \cite{zhang2019sagan,miyato2018spectral,brock2018biggan,liu2019funit,lim2017geometric} during 200K iterations. We employ exponential moving average of the generator \cite{karras2018pggan,yazici2018emagan,liu2019funit,karras2019stylegan}.
For the Thai-printing dataset, we use a learning rate of 0.00005 for the generator and 0.0001 for the discriminator with 250K training iterations while other settings are same as the Korean experiments.

\subsection{Evaluation classifier implementation details}

Two different ResNet-50~\cite{he2016_cvpr_resnet} are separately trained for the content and the style classifiers with Korean and Thai scripts. The classifiers are optimized using the Adam optimizer~\cite{kingma2015adam} with 20 epochs. We expect that more recent Adam variants, \eg, RAdam~\cite{radam} or AdamP~\cite{heo2020adamp}, further improve the classifier performances. The content classifier is supervised to predict a correct character, while the style classifier is trained to predict a font label. We randomly use 85\% of the data points as the train data and the remained data points are used for the validation. Unlike the \model{} training, this strategy shows all characters and fonts to classifiers. In our experiment, every classifier achieves over 96\% of validation accuracy: 97.9\% Korean content accuracy, 96.0\% Korean style accuracy, 99.6\% Thai content accuracy, and 99.99\% Thai style accuracy. Note that all classifiers are only used for the evaluation but not for the \model{} training.

\subsection{User study details}

30 different styles (fonts) from the Korean-unrefined dataset are selected for the user study. We randomly choose 8 characters for each style and generate the characters with four different methods: EMD~\cite{zhang2018_cvpr_emd}, AGIS-Net~\cite{gao2019agisnet}, FUNIT~\cite{liu2019funit}, and \model{} (ours). We also provide ground truth characters for the selected characters to the users for the comparison. Users chose the best method in 3 different criteria (content / style / preference). For each question, we randomly shuffle the methods to keep anonymity of methods. To sum up, we got 3,420 responses from 38 Korean natives with $30\times 3$ items.

\section{Additional Results}
\label{sec:appendix_results}
\subsection{Reference set sensitivity}
\import{tables/}{sampling-sensitivity.tex}

In all experiments, we select the few-shot samples randomly while satisfying the compositionality. Here, we show that the reference sample selection sensitivity of the proposed method. Table~\ref{table:sampling-sensitivity} shows the Korean-handwriting generation results of the eight different runs with different sample selections. The results support that \model{} is robust to the reference sample selection.

\import{tables/}{appendix-unrefined-quantitative.tex}

\subsection{Results on the Korean-unrefined dataset}

\begin{figure}[t]
    \centering
    \includegraphics[page=1,width=0.99\textwidth]{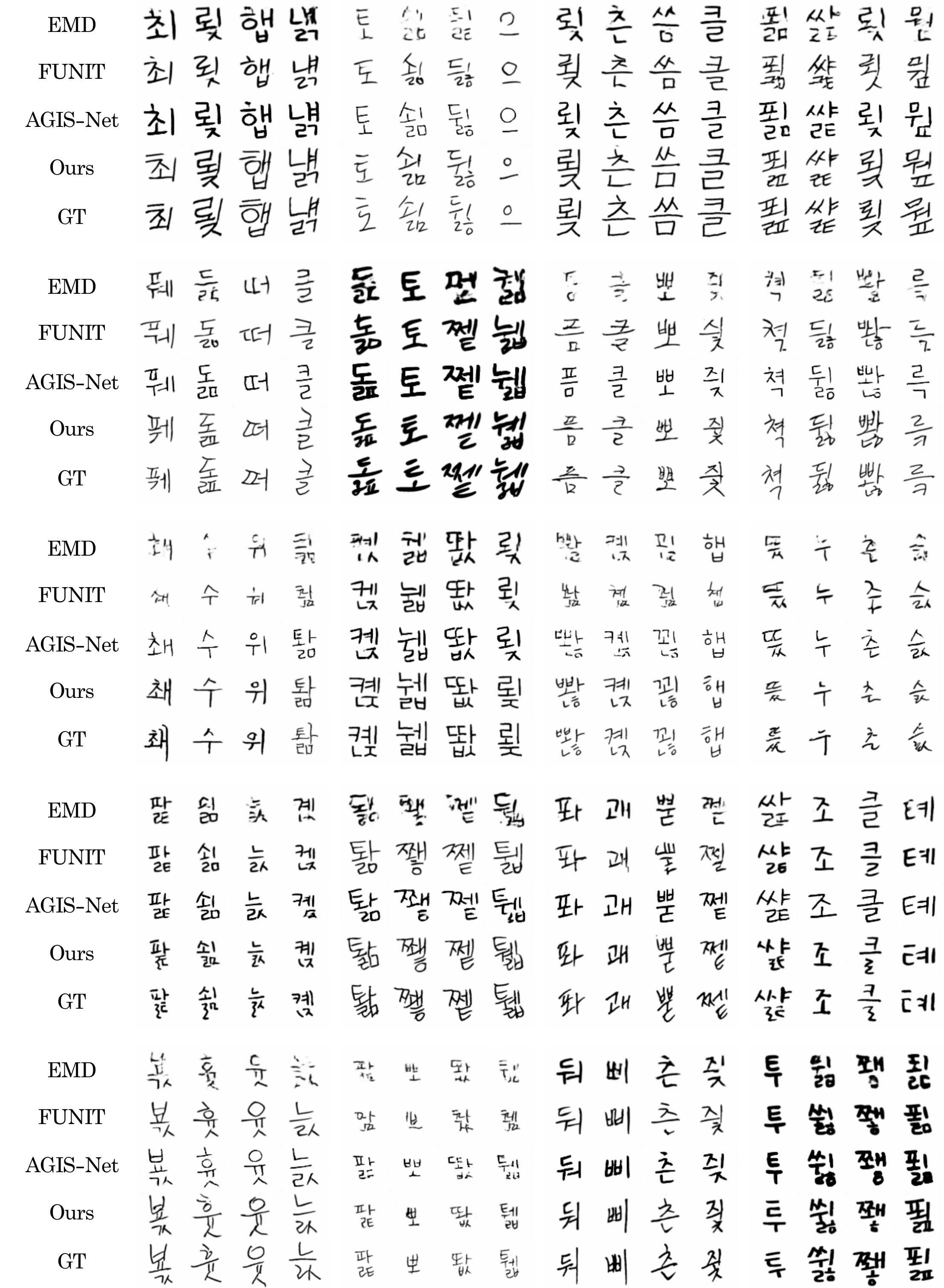}
    \caption{{\bf Samples for the user study.} The Korean-unrefined dataset is used.}
    \label{fig:user-study-more}
\end{figure}

Table \ref{table:quantitative-unrefined-userstudy} shows the quantitative evaluation results of the Korean-unrefined dataset used for the user study. We use the classifiers trained by the Korean-handwriting dataset for the evaluation. Hence, we only report the perceptual distance and mFID while accuracies are not measurable by the classifiers. In all evaluation metrics, \model{} consistently shows the remarkable performance as other datasets. The example visual samples are shown in Figure \ref{fig:user-study-more}.

\subsection{Ablation study}

Table \ref{table:ablation-full} shows the full ablation study results including all evaluation metrics. As the observations in the main manuscript, all metrics show similar behavior with the averaged accuracies; our proposed components and objective functions significantly improve the generation quality.

\import{tables/}{appendix-ablation-fulltable.tex}
\clearpage

\subsection{Failure cases}
\import{}{tables/failures.tex}

We illustrate the three failure types of our method in Figure~\ref{fig:failures}. First, \model{} can fail to generate the glyphs with the correct content due to the high complexity of the glyph. For example, some samples lose their contents -- See from the first to the third column of Figure~\ref{fig:failures} (a). In practice, developing a content failure detector and a user-guided font correction system can be a solution. Another widely-observed failure case caused by the multi-modality of components, \ie, a component can have multiple styles. Since our scenario assumes that a model only observes one sample for each component, the problem is often ill-posed. Similar ill-posedness problem is also occurred in the colorization problem, and usually addressed by a human-guided algorithm~\cite{zhang2017real}. Similarly, a user-guided font correction algorithm will be an interesting future research.

Finally, we report the cases caused by the errors in the ground truth samples. Note that the samples in Figure~\ref{fig:failures} are generated by the Korean-unrefined dataset which can include inherent errors. When the reference glyphs are damaged as the two rightmost samples in the figure, it is difficult to disentangle style and content from the reference set. Due to the strong compositionality regularization by the proposed dual memory architecture, our model tries to use the memorized local styles while ignoring the damaged reference style.

\subsection{Examples of various component shape in generated glyphs}
\import{}{tables/shape-varying.tex}
We provide more examples of the generated samples by \model{} with the same component in Figure~\ref{fig:shape-varying}. The figure shows that the shape of each component is varying by different sub-glyphs compositions as described in Figure~\ref{fig:compositional-script} of the main manuscript. Note that all components are observed a few times (usually once) as the reference. These observations support that our model does not simply copy the reference components, but can properly extract local styles and combine them with global composition information and intrinsic shape stored in persistent memory. To sum up, we conclude that \model{} disentangles local style and global composition information well, and generates the high quality font library with only a few references.

\subsection{Generalization gap between seen and unseen fonts}
\import{}{tables/generalization-gap.tex}

We provide additional benchmarking results on the seen fonts in Table~\ref{table:generalization-gap}. Note that Table~\ref{table:quantitative-kor} and \ref{table:quantitative-thai} in the main manuscript are measured in the unseen fonts only. Simply, ``seen fonts'' can be interpreted as the training performances, and ``unseen fonts'' as the validation performances. The comparison methods such as \emd{}, \funit{}, \agis{}, show remarkably good results on the training data (seen fonts) but fail to generalize the performance on the validation set (unseen fonts). We also report the generalization gap between the seen and unseen fonts in Table~\ref{table:generalization-gap}. The results show that comparison methods suffer from the style memorization issue, what we discussed in the nearest neighbor analysis, and cannot be generalizable to the unseen font styles. In contrast, our method shows significantly better generalization gap comparing to others.

\end{document}

%% file: tables/quantitative-new.tex
\begin{table}[t]
\centering
\caption{{\bf Quantatitive evaluation on the Korean-handwriting dataset.} We evaluate the methods on the seen and unseen character sets. Higher is better, except perceptual distance (PD) and mFID.}
\label{table:quantitative-kor}
\small
\setlength{\tabcolsep}{0.3em}
\begin{tabular}{@{}lcccccccccc@{}}
\toprule
         & \multicolumn{2}{c}{Pixel-level} && \multicolumn{3}{c}{Content-aware} && \multicolumn{3}{c}{Style-aware}                                      \\
         & SSIM & MS-SSIM && Acc(\%) & PD & mFID && Acc(\%) & PD & mFID \\ \midrule
\multicolumn{11}{c}{Evaluation on the {\bf seen} character set during training}\\ \midrule
\emd               & 0.691       & 0.361       && 80.4       & 0.084       & 138.2      && 5.1        & 0.089       & 134.4      \\
\funit             & 0.686       & 0.369       && 94.5       & 0.030       & 42.9       && 5.1        & 0.087       & 146.7      \\
\agis          & 0.694       & 0.399       && {\bf 98.7} & {\bf 0.018} & 23.9       && 8.2        & 0.088       & 141.1      \\
\ours   & {\bf 0.704} & {\bf 0.457} && 98.1       & {\bf 0.018} & {\bf 22.1} && {\bf 64.1} & {\bf 0.038} & {\bf 34.6} \\ 
\midrule
\multicolumn{11}{c}{Evaluation on the {\bf unseen} character set during training}\\ \midrule
\emd               & 0.696       & 0.362       && 76.4       & 0.095       & 155.3      && 5.2        & 0.089       & 139.6 \\
\funit             & 0.690       & 0.372       && 93.3       & 0.034       & 48.4       && 5.6        & 0.087       & 149.5 \\
\agis          & 0.699       & 0.398       && 98.3       & 0.019       & 25.9       && 7.5        & 0.089       & 146.1 \\
\model{} (ours)   & {\bf 0.707} & {\bf 0.455} && {\bf 98.5} & {\bf 0.018} & {\bf 20.8} && {\bf 62.6} & {\bf 0.039} & {\bf 40.5} \\ \bottomrule
\end{tabular}
\end{table}

\begin{table}[t]
\centering
\caption{{\bf Quantatitive evaluation on the Thai-printing dataset.} We evaluate the methods on the seen and unseen character sets. Higher is better, except perceptual distance (PD) and mFID.}
\label{table:quantitative-thai}
\small
\setlength{\tabcolsep}{0.3em}
\begin{tabular}{@{}lcccccccccc@{}}
\toprule
         & \multicolumn{2}{c}{Pixel-level} && \multicolumn{3}{c}{Content-aware} && \multicolumn{3}{c}{Style-aware}                                      \\
         & SSIM & MS-SSIM && Acc(\%) & PD & mFID && Acc(\%) & PD & mFID \\ \midrule
\multicolumn{11}{c}{Evaluation on the {\bf seen} character set during training}\\ \midrule
\emd & 0.773 & 0.640 &  & 86.3 & 0.115 & 215.4  &  & 3.2  & 0.087 & 172.0 \\
\funit & 0.712 & 0.449 &  & 45.8 & 0.566 & 1133.8 &  & 4.6  & 0.084 & 167.9 \\
\agis & 0.758 & 0.624 &  & {\bf 87.2} & {\bf 0.091} & {\bf 165.2}  &  & 15.5 & 0.074 & 145.2 \\
\ours     & {\bf 0.776} & {\bf 0.697} && 87.0 & 0.103 & 198.7  && {\bf 50.3} & {\bf 0.037} & {\bf 69.4} \\
\midrule
\multicolumn{11}{c}{Evaluation on the {\bf unseen} character set during training}\\ \midrule
\emd & 0.770 & 0.636 &  & 85.0 & 0.123 & 231.0  &  & 3.4  & 0.087 & 171.6 \\
\funit & 0.708 & 0.442 &  & 45.0 & 0.574 & 1149.8 &  & 4.7  & 0.084 & 166.9 \\
\agis & 0.755 & 0.618 &  & 85.4 & 0.103 & {\bf 188.4}  &  & 15.8 & 0.074 & 145.1 \\
\ours & {\bf 0.773} & {\bf 0.693} &  & {\bf 87.2} & {\bf 0.101} & 195.9  &  & {\bf 50.6} & {\bf 0.037} & {\bf 69.6} \\ \bottomrule
\end{tabular}
\end{table}

%% file: tables/user_study.tex

\begin{table}[t]
\centering
\caption{{\bf User study results on the Korean-unrefined dataset.} Each number is the preferred model output out of $3,420$ responses.}
\label{table:user_study}
\small
\begin{tabular}{@{}lcccc@{}}
\toprule
                            & \emd & \funit & \agis & \ours          \\ \midrule
Best content preserving   & 1.33\% & 9.17\% & {\bf48.67\%} & 40.83\%       \\
Best stylization          & 1.71\% & 8.14\% & 17.44\%      & {\bf 72.71\%} \\
Most preferred              & 1.23\% & 9.74\% & 16.40\%      & {\bf 72.63\%} \\ \bottomrule
\end{tabular}
\end{table}

%% file: tables/ablation.tex
\begin{table}[t!]
\caption{{\bf Ablation studies on the Korean-handwriting dataset.} Each content and style score is an average of the seen and unseen accuracies. Hmean denotes the harmonic mean of content and style scores.} \label{table:ablation}
\begin{subtable}[t]{0.54\textwidth}
\centering
\caption{Impact of the memory modules.}
\label{table:ablation-components}
\small
\begin{tabular}{@{}lccc@{}}
\toprule
                      & Content & Style & Hmean \\ \midrule
Baseline              & 96.6    & 6.5   & 12.2  \\
$+$ Dynamic memory   & {\bf 99.8}    & 32.0  & 48.5  \\
$+$ Persistent memory & 97.6    & 46.2  & 62.8  \\
$+$ Compositional $G$ & 98.3    & {\bf 63.3}  & {\bf 77.0}  \\ \bottomrule
\end{tabular}
\end{subtable}
\hfill
\begin{subtable}[t]{0.45\textwidth}
\centering
\caption{Impact of the objective functions.}
\label{table:ablation-objective}
\small
\begin{tabular}{@{}lccc@{}}
\toprule
                      & Content & Style & Hmean \\ \midrule
Full                  & {\bf 98.3} & {\bf 63.3} & {\bf 77.0} \\
Full $- \,\mathcal L_{l1}$   & 97.3 & 53.8 & 69.3 \\
Full $- \,\mathcal L_{feat}$ & 97.8 & 51.3 & 67.3 \\
Full $- \,\mathcal L_{cls}$  & 3.1  & 16.0 & 5.2  \\ \bottomrule
\end{tabular}
\end{subtable}
\end{table}

%% file: tables/vis_nn_l1.tex
\begin{figure}[t]
\centering
\setlength{\tabcolsep}{2pt}
\renewcommand{\arraystretch}{0.1}
\scriptsize
\begin{tabular}{c|cc|cc|cc|cc}
\includegraphics[align=c, width=0.07\textwidth]{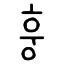} &
\includegraphics[align=c, width=0.07\textwidth]{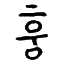} & 
\includegraphics[align=c, width=0.07\textwidth]{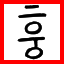} &
\includegraphics[align=c, width=0.07\textwidth]{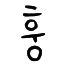} & 
\includegraphics[align=c, width=0.07\textwidth]{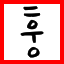} &
\includegraphics[align=c, width=0.07\textwidth]{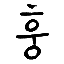} &
\includegraphics[align=c, width=0.07\textwidth]{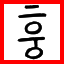} &
\includegraphics[align=c, width=0.07\textwidth]{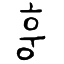} &
\includegraphics[align=c, width=0.07\textwidth]{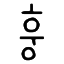}
\\
&&&&&&&&
\\
\includegraphics[align=c, width=0.07\textwidth]{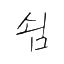} &
\includegraphics[align=c, width=0.07\textwidth]{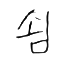} & \includegraphics[align=c, width=0.07\textwidth]{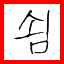} &
\includegraphics[align=c, width=0.07\textwidth]{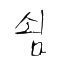} & \includegraphics[align=c, width=0.07\textwidth]{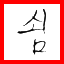} &
\includegraphics[align=c, width=0.07\textwidth]{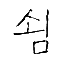} &
\includegraphics[align=c, width=0.07\textwidth]{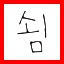} &
\includegraphics[align=c, width=0.07\textwidth]{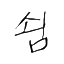} &
\includegraphics[align=c, width=0.07\textwidth]{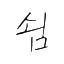}
\\
&&&&&&&&
\\
\includegraphics[align=c, width=0.07\textwidth]{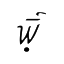} &
\includegraphics[align=c, width=0.07\textwidth]{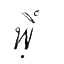} & \includegraphics[align=c, width=0.07\textwidth]{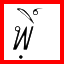} &
\includegraphics[align=c, width=0.07\textwidth]{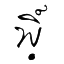} & \includegraphics[align=c, width=0.07\textwidth]{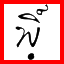} &
\includegraphics[align=c, width=0.07\textwidth]{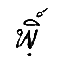} &
\includegraphics[align=c, width=0.07\textwidth]{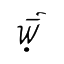} &
\includegraphics[align=c, width=0.07\textwidth]{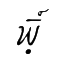} &
\includegraphics[align=c, width=0.07\textwidth]{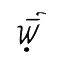}
\\
&&&&&&&&
\\
GT & EMD & EMD & FUNIT & FUNIT & AGIS-Net & AGIS-Net & Ours & Ours \\
&&&&&&&&
\\
 & (output) & (NN) & (output) & (NN) & (output) & (NN) & (output) & (NN) \\
\end{tabular}
\caption{{\bf Nearest neighbor analysis.} We report the generated images by each model (output) with the given unseen reference style (GT) and the ground truth samples whose label is predicted by the style classifier (NN). Red boxed samples denote training samples. We can conclude that the baseline methods are overfitted to the training style while ours easily generalizes to unseen style.}
\label{fig:style-memorization}
\end{figure}

%% file: tables/sampling-sensitivity.tex
\begin{table}[t]
\centering
\caption{{\bf Reference sample sensitivity.} Eight different runs of \model{} with different reference samples in the Korean-handwriting dataset.}

\label{table:sampling-sensitivity}
\small
\setlength{\tabcolsep}{0.3em}
\begin{tabular}{@{}lcccccccccc@{}}
\toprule
         & \multicolumn{2}{c}{Pixel-level} && \multicolumn{3}{c}{Content-aware} && \multicolumn{3}{c}{Style-aware}                                      \\
         & SSIM & MS-SSIM && Acc(\%) & PD & mFID && Acc(\%) & PD & mFID \\ 
          \midrule
Run 1   & 0.704         & 0.457           &  & 98.1\%        & 0.018 & 22.1 &  & 64.1\%      & 0.038 & 34.6 \\ 
Run 2   & 0.702         & 0.452           &  & 98.8\%        & 0.016 & 19.9 &  & 64.2\%      & 0.038 & 37.2 \\
Run 3   & 0.701         & 0.456           &  & 98.0\%        & 0.018 & 23.4 &  & 66.0\%      & 0.037 & 35.2 \\
Run 4   & 0.702         & 0.451           &  & 97.8\%        & 0.019 & 22.9 &  & 65.0\%      & 0.038 & 36.7 \\
Run 5   & 0.701         & 0.453           &  & 98.2\%        & 0.018 & 22.9 &  & 64.8\%      & 0.038 & 36.4 \\
Run 6   & 0.703         & 0.460           &  & 97.2\%        & 0.020 & 24.8 &  & 67.8\%      & 0.036 & 34.0 \\
Run 7   & 0.700         & 0.447           &  & 98.3\%        & 0.018 & 21.9 &  & 64.8\%      & 0.037 & 36.6 \\
Run 8   & 0.701         & 0.451           &  & 98.2\%        & 0.018 & 22.2 &  & 65.8\%      & 0.037 & 35.4 \\
\midrule
Avg. & 0.702         & 0.453           &  & 98.1\%        & 0.018 & 22.5 &  & 65.3\%      & 0.037 & 35.8 \\
Std. & 0.001         & 0.004           &  & 0.4\%         & 0.001 & 1.4  &  & 1.2\%       & 0.001 & 1.1 \\
\bottomrule
\end{tabular}
\end{table}

%% file: tables/appendix-unrefined-quantitative.tex
\begin{table}[h!]
\centering
\caption{{\bf Quantatitive Evaluation on the Korean-unrefined dataset.} Higher is better, except perceptual distance (PD) and mFID.}
\label{table:quantitative-unrefined-userstudy}
\small
\setlength{\tabcolsep}{0.3em}
\begin{tabular}{@{}lcccccccc@{}}
\toprule
         & \multicolumn{2}{c}{Pixel-level} && \multicolumn{2}{c}{Content-aware} && \multicolumn{2}{c}{Style-aware}                                      \\
         & SSIM & MS-SSIM && PD & mFID && PD & mFID \\ \midrule

\emd & 0.716 & 0.340 && 0.106 & 99.2 && 0.079 & 93.3 \\ 
\funit & 0.711 & 0.311 && 0.080 & 87.0 && 0.066 & 79.4 \\ 
\agis & 0.708 & 0.334 && 0.052 & 67.2 && 0.089 & 134.5 \\ 
\ours & \textbf{0.726} & \textbf{0.387} && \textbf{0.048} & \textbf{46.2} && \textbf{0.046} & \textbf{31.5} \\
\bottomrule
\end{tabular}
\end{table}

%% file: tables/appendix-ablation-fulltable.tex
\begin{table}[t]
\centering
\caption{{\bf Ablation studies on the Korean-handwriting dataset.} Higher is better, except perceptual distance (PD) and mFID.}
\label{table:ablation-full}
\small
\setlength{\tabcolsep}{0.3em}

\begin{subtable}[t]{\textwidth}
    \caption{Impact of components. DM, PM, and Comp. $G$ denote dynamic memory, persistent memory, and compositional generator, respectively.}
    \label{table:ablation-full-component}
    \centering
    \small
    \begin{tabular}{@{}lcccccccccc@{}}
    \toprule
             & \multicolumn{2}{c}{Pixel-level} && \multicolumn{3}{c}{Content-aware} && \multicolumn{3}{c}{Style-aware}                                      \\
             & SSIM & MS-SSIM && Acc(\%) & PD & mFID && Acc(\%) & PD & mFID \\ \midrule
    \multicolumn{11}{c}{Evaluation on the {\bf seen} character set during training}\\ \midrule
    Baseline              & 0.689 & 0.373 && 96.7 & 0.026 & 33.6 && 6.5 & 0.084 & 132.7 \\ 
    + DM      & 0.702 & 0.424 && \textbf{99.7} & \textbf{0.015} & \textbf{19.5} && 31.8 & 0.060 & 77.6 \\ 
    + PM   & \textbf{0.704} & 0.435 && 97.7 & 0.020 & 26.9 && 46.6 & 0.049 & 57.1 \\ 
    + Comp. $G$ & \textbf{0.704} & \textbf{0.457} && 98.1 & 0.018 & 22.1 && \textbf{64.1} & \textbf{0.038} & \textbf{34.6} \\ 
    \midrule
    \multicolumn{11}{c}{Evaluation on the {\bf unseen} character set during training}\\ \midrule
    Baseline              & 0.693 & 0.375 && 96.6 & 0.027 & 34.3 && 6.5 & 0.084 & 134.8 \\ 
    + DM      & 0.705 & 0.423 && \textbf{99.8} & \textbf{0.015} & \textbf{19.5} && 32.3 & 0.060 & 81.0 \\ 
    + PM   & \textbf{0.707} & 0.432 && 97.6 & 0.022 & 28.9 && 45.9 & 0.050 & 61.4 \\ 
    + Comp. $G$ & \textbf{0.707} & \textbf{0.455} && 98.5 & 0.018 & 20.8 && \textbf{62.6} & \textbf{0.039} & \textbf{40.5} \\
    \bottomrule
    \end{tabular}
\end{subtable}
    
\begin{subtable}[t]{\textwidth}
    \caption{Impact of objective functions.}
    \label{table:ablation-full-objective}
    \centering
    \small
    \begin{tabular}{@{}lcccccccccc@{}}
    \toprule
             & \multicolumn{2}{c}{Pixel-level} && \multicolumn{3}{c}{Content-aware} && \multicolumn{3}{c}{Style-aware}                                      \\
             & SSIM & MS-SSIM && Acc(\%) & PD & mFID && Acc(\%) & PD & mFID \\ \midrule
    \multicolumn{11}{c}{Evaluation on the {\bf seen} character set during training}\\ \midrule
    Full & \textbf{0.704} & \textbf{0.457} && \textbf{98.1} & \textbf{0.018} & \textbf{22.1} && \textbf{64.1} & \textbf{0.038} & \textbf{34.6} \\ 
    Full $- \mathcal L_{l1}$ & 0.695 & 0.407 && 97.0 & 0.022 & 27.9 && 53.4 & 0.046 & 48.3 \\ 
    Full $- \mathcal L_{feat}$ & 0.699 & 0.427 && 97.8 & 0.020 & 23.8 && 51.4 & 0.047 & 51.4 \\ 
    Full $- \mathcal L_{cls}$ & 0.634 & 0.223 && 3.0 & 0.488 & 965.3 && 16.2 & 0.082 & 118.9 \\ 
    \midrule
    \multicolumn{11}{c}{Evaluation on the {\bf unseen} character set during training}\\ \midrule
    Full & \textbf{0.707} & \textbf{0.455} && \textbf{98.5} & \textbf{0.018} & \textbf{20.8} && \textbf{62.6} & \textbf{0.039} & \textbf{40.5} \\ 
    Full $- \mathcal L_{l1}$ & 0.697 & 0.401 && 97.5 & 0.023 & 26.8 && 54.3 & 0.046 & 52.3 \\ 
    Full $- \mathcal L_{feat}$ & 0.701 & 0.423 && 97.8 & 0.020 & 24.1 && 51.2 & 0.048 & 56.0 \\ 
    Full $- \mathcal L_{cls}$ & 0.636 & 0.220 && 3.2 & 0.486 & 960.7 && 15.9 & 0.082 & 123.7 \\
    \bottomrule
    \end{tabular}
\end{subtable}
\end{table}


%% file: tables/failures.tex
\newcommand{\vcentered}[1]{\begin{tabular}{l} #1 \end{tabular}}

\begin{figure}[t]
\centering
\setlength{\tabcolsep}{3pt}
\renewcommand{\arraystretch}{0.4}
\begin{tabular}{c|c|c|c}
\vcentered{GT} &
\includegraphics[align=c, width=0.27\textwidth]{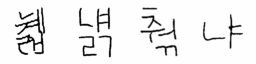} &
\includegraphics[align=c, width=0.27\textwidth]{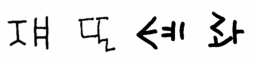} &
\includegraphics[align=c, width=0.27\textwidth]{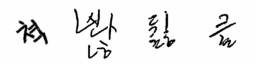}
\\
\vcentered{Ours} &
\includegraphics[align=c, width=0.27\textwidth]{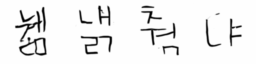} &
\includegraphics[align=c, width=0.27\textwidth]{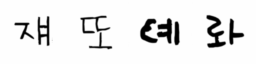} &
\includegraphics[align=c, width=0.27\textwidth]{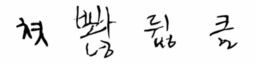}
\\
\multicolumn{1}{c}{} & \multicolumn{1}{c}{} & \multicolumn{1}{c}{} & \\
\multicolumn{1}{c}{} & \multicolumn{1}{c}{(a) content failure} & \multicolumn{1}{c}{(b) style failure} & (c) content and style \\
\multicolumn{1}{c}{} & \multicolumn{1}{c}{} & \multicolumn{1}{c}{} & failure \\
\end{tabular}
\caption{{\bf Failure cases.} Examples of generated samples by \model{} with incorrect content or insufficient stylization.}
\label{fig:failures}
\end{figure}

%% file: tables/shape-varying.tex
\begin{figure}[t]
\centering
\setlength{\tabcolsep}{3pt}
\renewcommand{\arraystretch}{0.4}
\begin{tabular}{c:c:c}
\includegraphics[align=c, width=0.3\textwidth]{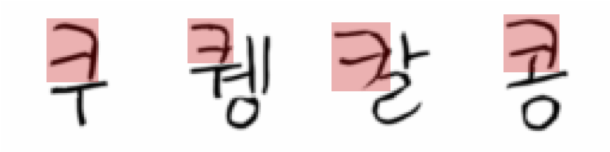} &
\includegraphics[align=c, width=0.3\textwidth]{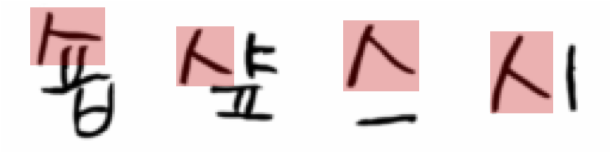} &
\includegraphics[align=c, width=0.3\textwidth]{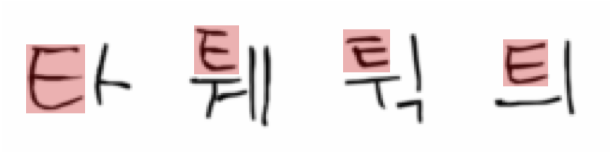}
\\ \hdashline
\includegraphics[align=c, width=0.3\textwidth]{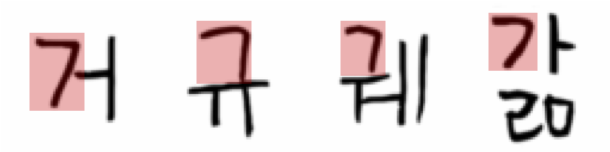} &
\includegraphics[align=c, width=0.3\textwidth]{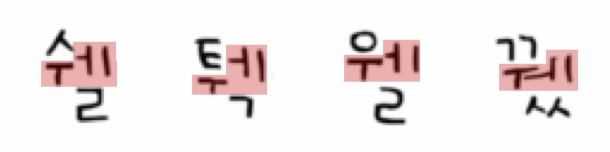} &
\includegraphics[align=c, width=0.3\textwidth]{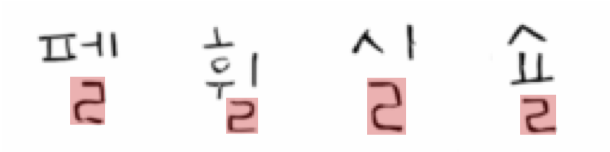}
\\
\end{tabular}
\caption{{\bf Varying shape of single component.} Six visual examples show the variety of a component (with red boxes) across different characters. The results show that \model{} generates samples with various component shapes by the compositionality.}
\label{fig:shape-varying}
\end{figure}

%% file: tables/generalization-gap.tex
\begin{table}[t]
\centering
\caption{{\bf Style generalization gap on the Korean-handwriting dataset.} We compute the differences of style-aware scores between seen and unseen font sets. The evaluation uses unseen character set. Smaller gap indicates better generalization.}
\label{table:generalization-gap}
\small
\setlength{\tabcolsep}{0.3em}
\begin{tabular}{@{}lccccccccccc@{}}
\toprule
         & \multicolumn{3}{c}{Seen fonts} && \multicolumn{3}{c}{Unseen fonts} && \multicolumn{3}{c}{Gap}                                      \\
        & Acc(\%)        & PD             & mFID         &  & Acc(\%)       & PD             & mFID          &  & Acc(\%)       & PD             & mFID \\ \midrule
\multicolumn{12}{c}{Evaluation on the \textbf{Korean-handwriting} dataset}\\ \midrule
EMD    & 74.0          & 0.032          & 31.9         &  & 5.2           & 0.089          & 139.6         &  & 68.9          & 0.057          & 107.8         \\
FUNIT  & \textbf{98.6} & \textbf{0.015} & \textbf{8.3} &  & 5.6           & 0.087          & 149.5         &  & 93.0          & 0.072          & 141.2         \\
AGIS-Net   & 95.8          & 0.018          & 13.4         &  & 7.5           & 0.089          & 146.1         &  & 88.3          & 0.071          & 132.7         \\
DM-Font   & 82.1          & 0.026          & 16.9         &  & \textbf{62.6} & \textbf{0.039} & \textbf{40.5} &  & \textbf{19.5} & \textbf{0.013} & \textbf{23.6} \\
\midrule
\multicolumn{12}{c}{Evaluation on the \textbf{Thai-printing} dataset}\\ \midrule
EMD    & \textbf{99.5} & \textbf{0.001} & \textbf{1.0} &  & 3.4           & 0.087          & 171.6         &  & 96.1          & 0.086          & 170.6         \\
FUNIT  & 97.0          & 0.004          & 5.0          &  & 4.7           & 0.084          & 166.9         &  & 92.4          & 0.080          & 161.9         \\
AGIS-Net   & 84.6          & 0.016          & 28.6         &  & 15.8          & 0.074          & 145.1         &  & 68.8          & 0.058          & 116.5         \\
DM-Font   & 90.2          & 0.009          & 13.5         &  & \textbf{50.6} & \textbf{0.037} & \textbf{69.6} &  & \textbf{39.6} & \textbf{0.029} & \textbf{56.1} \\
\bottomrule
\end{tabular}
\end{table}

%% file: eccv2020arxiv.bbl
\begin{thebibliography}{10}
\providecommand{\url}[1]{\texttt{#1}}
\providecommand{\urlprefix}{URL }
\providecommand{\doi}[1]{https://doi.org/#1}

\bibitem{zi2zi}
zi2zi: Master chinese calligraphy with conditional adversarial networks,
  \url{https://github.com/kaonashi-tyc/zi2zi}

\bibitem{azadi2018mcgan}
Azadi, S., Fisher, M., Kim, V.G., Wang, Z., Shechtman, E., Darrell, T.:
  Multi-content gan for few-shot font style transfer. In: IEEE Conference on
  Computer Vision and Pattern Recognition (2018)

\bibitem{bello2019_ICCV_AttentionAugmentedConv}
Bello, I., Zoph, B., Vaswani, A., Shlens, J., Le, Q.V.: Attention augmented
  convolutional networks. In: Proceedings of the IEEE International Conference
  on Computer Vision. pp. 3286--3295 (2019)

\bibitem{brock2018biggan}
Brock, A., Donahue, J., Simonyan, K.: Large scale {GAN} training for high
  fidelity natural image synthesis. In: International Conference on Learning
  Representations (2019)

\bibitem{cao2019gcnet}
Cao, Y., Xu, J., Lin, S., Wei, F., Hu, H.: {GCNet}: Non-local networks meet
  squeeze-excitation networks and beyond. In: IEEE International Conference on
  Computer Vision Workshops (2019)

\bibitem{chang2018_wacv_densecyclegan}
Chang, B., Zhang, Q., Pan, S., Meng, L.: Generating handwritten chinese
  characters using cyclegan. In: IEEE Winter Conference on Applications of
  Computer Vision (2018)

\bibitem{chang2018_bmvc_hgan}
Chang, J., Gu, Y., Zhang, Y., Wang, Y.F.: Chinese handwriting imitation with
  hierarchical generative adversarial network. In: British Machine Vision
  Conference (2018)

\bibitem{stargan}
Choi, Y., Choi, M., Kim, M., Ha, J.W., Kim, S., Choo, J.: {StarGAN}: Unified
  generative adversarial networks for multi-domain image-to-image translation.
  In: IEEE Conference on Computer Vision and Pattern Recognition (2018)

\bibitem{starganv2}
Choi, Y., Uh, Y., Yoo, J., Ha, J.W.: {StarGAN} v2: Diverse image synthesis for
  multiple domains. In: IEEE Conference on Computer Vision and Pattern
  Recognition (2020)

\bibitem{gao2019agisnet}
Gao, Y., Guo, Y., Lian, Z., Tang, Y., Xiao, J.: Artistic glyph image synthesis
  via one-stage few-shot learning. ACM Transactions on Graphics  (2019)

\bibitem{gatys2016neuralstyle}
Gatys, L.A., Ecker, A.S., Bethge, M.: Image style transfer using convolutional
  neural networks. In: IEEE Conference on Computer Vision and Pattern
  Recognition (2016)

\bibitem{gan}
Goodfellow, I., Pouget-Abadie, J., Mirza, M., Xu, B., Warde-Farley, D., Ozair,
  S., Courville, A., Bengio, Y.: Generative adversarial nets. In: Advances in
  neural information processing systems (2014)

\bibitem{koreantextbook}
Han, J., Lee, Y., Ahn, S.: Korean font design textbook. Ahn graphics (2009)

\bibitem{he2016_cvpr_resnet}
He, K., Zhang, X., Ren, S., Sun, J.: Deep residual learning for image
  recognition. In: IEEE Conference on Computer Vision and Pattern Recognition
  (2016)

\bibitem{heo2020adamp}
Heo, B., Chun, S., Oh, S.J., Han, D., Yun, S., Uh, Y., Ha, J.W.: Slowing down
  the weight norm increase in momentum-based optimizers. arXiv preprint
  arXiv:2006.08217  (2020)

\bibitem{heusel2017_nips_ttur_fid}
Heusel, M., Ramsauer, H., Unterthiner, T., Nessler, B., Hochreiter, S.: {GAN}s
  trained by a two time-scale update rule converge to a local nash equilibrium.
  In: Advances in Neural Information Processing Systems (2017)

\bibitem{adain}
Huang, X., Belongie, S.J.: Arbitrary style transfer in real-time with adaptive
  instance normalization. In: IEEE International Conference on Computer Vision
  (2017)

\bibitem{isola2017_cvpr_pix2pix}
Isola, P., Zhu, J.Y., Zhou, T., Efros, A.A.: Image-to-image translation with
  conditional adversarial networks. In: IEEE Conference on Computer Vision and
  Pattern Recognition (2017)

\bibitem{jiang2017dcfont}
Jiang, Y., Lian, Z., Tang, Y., Xiao, J.: {DCFont}: An end-to-end deep chinese
  font generation system. In: SIGGRAPH Asia (2017)

\bibitem{jiang2019_aaai_scfont}
Jiang, Y., Lian, Z., Tang, Y., Xiao, J.: {SCFont}: Structure-guided chinese
  font generation via deep stacked networks. In: AAAI Conference on Artificial
  Intelligence (2019)

\bibitem{johnson2016_eccv_perceptual}
Johnson, J., Alahi, A., Fei-Fei, L.: Perceptual losses for real-time style
  transfer and super-resolution. In: European Conference on Computer Vision
  (2016)

\bibitem{karras2018pggan}
Karras, T., Aila, T., Laine, S., Lehtinen, J.: Progressive growing of {GAN}s
  for improved quality, stability, and variation. In: International Conference
  on Learning Representations (2018)

\bibitem{karras2019stylegan}
Karras, T., Laine, S., Aila, T.: A style-based generator architecture for
  generative adversarial networks. In: IEEE Conference on Computer Vision and
  Pattern Recognition (2019)

\bibitem{kingma2015adam}
Kingma, D.P., Ba, J.: Adam: A method for stochastic optimization. In:
  International Conference on Learning Representations (2015)

\bibitem{wct}
Li, Y., Fang, C., Yang, J., Wang, Z., Lu, X., Yang, M.H.: Universal style
  transfer via feature transforms. In: Advances in Neural Information
  Processing Systems (2017)

\bibitem{photowct}
Li, Y., Liu, M.Y., Li, X., Yang, M.H., Kautz, J.: A closed-form solution to
  photorealistic image stylization. In: European Conference on Computer Vision
  (2018)

\bibitem{lim2017geometric}
Lim, J.H., Ye, J.C.: Geometric gan. arXiv preprint arXiv:1705.02894  (2017)

\bibitem{Lin_2017_CVPR_FPN}
Lin, T.Y., Dollar, P., Girshick, R., He, K., Hariharan, B., Belongie, S.:
  Feature pyramid networks for object detection. In: IEEE Conference on
  Computer Vision and Pattern Recognition (2017)

\bibitem{radam}
Liu, L., Jiang, H., He, P., Chen, W., Liu, X., Gao, J., Han, J.: On the
  variance of the adaptive learning rate and beyond. In: International
  Conference on Learning Representations (2020)

\bibitem{liu2019funit}
Liu, M.Y., Huang, X., Mallya, A., Karras, T., Aila, T., Lehtinen, J., Kautz,
  J.: Few-shot unsupervised image-to-image translation. In: IEEE International
  Conference on Computer Vision (2019)

\bibitem{celeba}
Liu, Z., Luo, P., Wang, X., Tang, X.: Deep learning face attributes in the
  wild. In: Proceedings of International Conference on Computer Vision (2015)

\bibitem{deepphotostyle}
Luan, F., Paris, S., Shechtman, E., Bala, K.: Deep photo style transfer. In:
  IEEE Conference on Computer Vision and Pattern Recognition (2017)

\bibitem{lyu2017_icdar_aegg}
Lyu, P., Bai, X., Yao, C., Zhu, Z., Huang, T., Liu, W.: Auto-encoder guided gan
  for chinese calligraphy synthesis. In: International Conference on Document
  Analysis and Recognition (2017)

\bibitem{mescheder2018training}
Mescheder, L., Geiger, A., Nowozin, S.: Which training methods for gans do
  actually converge? In: International Conference on Machine Learning (2018)

\bibitem{miyato2018spectral}
Miyato, T., Kataoka, T., Koyama, M., Yoshida, Y.: Spectral normalization for
  generative adversarial networks. In: International Conference on Learning
  Representations (2018)

\bibitem{ferjad2020ganeval}
Naeem, M.F., Oh, S.J., Uh, Y., Choi, Y., Yoo, J.: Reliable fidelity and
  diversity metrics for generative models. In: International Conference on
  Machine Learning (2020)

\bibitem{newell2016hourglass}
Newell, A., Yang, K., Deng, J.: Stacked hourglass networks for human pose
  estimation. In: European Conference on Computer Vision (2016)

\bibitem{srivatsan2019_emnlp_deepfactorization}
Srivatsan, N., Barron, J., Klein, D., Berg-Kirkpatrick, T.: A deep
  factorization of style and structure in fonts. In: Conference on Empirical
  Methods in Natural Language Processing (2019)

\bibitem{sun2018_ijcai_savae}
Sun, D., Ren, T., Li, C., Su, H., Zhu, J.: Learning to write stylized chinese
  characters by reading a handful of examples. In: International Joint
  Conference on Artificial Intelligence (2018)

\bibitem{vaswani2017_NIPS_Transformer_Attention}
Vaswani, A., Shazeer, N., Parmar, N., Uszkoreit, J., Jones, L., Gomez, A.N.,
  Kaiser, {\L}., Polosukhin, I.: Attention is all you need. In: Advances in
  neural information processing systems. pp. 5998--6008 (2017)

\bibitem{yazici2018emagan}
Yaz{\i}c{\i}, Y., Foo, C.S., Winkler, S., Yap, K.H., Piliouras, G.,
  Chandrasekhar, V.: The unusual effectiveness of averaging in {GAN} training.
  In: International Conference on Learning Representations (2019)

\bibitem{wct2}
Yoo, J., Uh, Y., Chun, S., Kang, B., Ha, J.W.: Photorealistic style transfer
  via wavelet transforms. In: International Conference on Computer Vision
  (2019)

\bibitem{zhang2019sagan}
Zhang, H., Goodfellow, I., Metaxas, D., Odena, A.: Self-attention generative
  adversarial networks. In: International Conference on Machine Learning (2019)

\bibitem{zhang2018_cvpr_lpips}
Zhang, R., Isola, P., Efros, A.A., Shechtman, E., Wang, O.: The unreasonable
  effectiveness of deep features as a perceptual metric. In: IEEE Conference on
  Computer Vision and Pattern Recognition (2018)

\bibitem{zhang2017real}
Zhang, R., Zhu, J.Y., Isola, P., Geng, X., Lin, A.S., Yu, T., Efros, A.A.:
  Real-time user-guided image colorization with learned deep priors. ACM
  Transactions on Graphics  (2017)

\bibitem{zhang2018_cvpr_emd}
Zhang, Y., Zhang, Y., Cai, W.: Separating style and content for generalized
  style transfer. In: IEEE Conference on Computer Vision and Pattern
  Recognition (2018)

\bibitem{zhu2017_iccv_cyclegan}
Zhu, J.Y., Park, T., Isola, P., Efros, A.A.: Unpaired image-to-image
  translation using cycle-consistent adversarial networks. In: IEEE
  International Conference on Computer Vision (2017)

\end{thebibliography}
